\documentclass[letterpaper]{article} 
\usepackage{aaai2027}  
\usepackage[hyphens]{url}  
\usepackage{graphicx} 
\usepackage{natbib}  
\usepackage{caption} 
\usepackage{algorithm}
\usepackage{algorithmic}
\usepackage{amsmath,amssymb,amsfonts,subcaption,multirow}
\usepackage{colortbl,xcolor,array}

\definecolor{strongfill}{RGB}{198,232,206}
\definecolor{strongtext}{RGB}{20,90,48}
\definecolor{modfill}{RGB}{230,244,233}
\definecolor{modtext}{RGB}{60,130,85}
\definecolor{neutfill}{RGB}{237,237,237}
\definecolor{neuttext}{RGB}{95,95,95}

\newcommand{\bst}[1]{\colorbox{strongfill}{\textcolor{strongtext}{\bfseries\fontsize{6.3}{6.3}\selectfont #1}}}
\newcommand{\bmo}[1]{\colorbox{modfill}{\textcolor{modtext}{\bfseries\fontsize{6.3}{6.3}\selectfont #1}}}
\newcommand{\bnt}[1]{\colorbox{neutfill}{\textcolor{neuttext}{\bfseries\fontsize{6.3}{6.3}\selectfont #1}}}

\usepackage{newfloat}
\usepackage{listings}
\DeclareCaptionStyle{ruled}{labelfont=normalfont,labelsep=colon,strut=off} 
\floatstyle{ruled}
\newfloat{listing}{tb}{lst}{}
\floatname{listing}{Listing}

\usepackage{booktabs}
\usepackage{caption}
\title{Future Confidence Distillation in Large Language Models}
\author{
    Sahil Kale\textsuperscript{\rm 1}
}
\affiliations{
    \textsuperscript{\rm 1}University of California, Los Angeles

    Los Angeles, USA\\
    sahilrkale@cs.ucla.edu
}

\begin{document}

\maketitle

\begin{abstract}

Reliable confidence estimation is essential for deploying large language models (LLMs) in confidence-aware systems, where downstream decisions such as retrieval, tool use, and adaptive computation depend on accurately estimating answer reliability. Existing approaches, however, largely treat confidence as a property of completed responses, overlooking how confidence-related information evolves throughout the answering process. In this work, we investigate confidence from a temporal perspective by comparing pre-solution Feeling-of-Knowing (FOK) and post-solution Judgement-of-Learning (JOL) confidence estimates across frontier and open-source LLMs. We show that post-solution confidence is consistently better calibrated and more discriminative than pre-solution confidence, while linear probes trained on hidden representations recover substantially richer confidence-related information than models explicitly verbalise. Building on this observation, we introduce \emph{future confidence distillation}, which trains predictors operating on pre-solution hidden representations using teacher confidence estimates produced by post-solution correctness probes. Despite requiring only pre-solution representations for inference, distilled predictors recover much of the calibration improvement achieved by post-solution confidence, remain highly sample efficient, and transfer across datasets within the same domain. Together, our findings demonstrate that confidence-related information evolves throughout the answering process and can be anticipated before answer generation is complete, enabling significantly more reliable yet low-cost confidence estimation.

\end{abstract}


\section{Introduction}

Large language models (LLMs) are increasingly deployed in high-stakes settings including question answering, reasoning systems, retrieval-augmented generation (RAG), autonomous agents, and decision-support pipelines \cite{li-etal-2024-fundamental,cheng-etal-2025-realm}. In such systems, models or downstream components often rely on internal confidence estimates to determine whether additional retrieval should be triggered, external tools should be invoked, more computation should be allocated, or a response should be deferred altogether \cite{li-etal-2025-metacognition}. Such a capability is often framed as \textit{self-knowledge} \cite{kale-vrn-2025-line} or \textit{metacognition} \cite{ma-etal-2025-llm-metacognition}. Consequently, the quality of a model's self-knowledge has become a critical factor of reliable, effective, and cost-efficient confidence-aware decision policies \cite{kale2025knowrlteachinglanguagemodels}.

A growing body of work studies self-knowledge via confidence estimation and calibration in LLMs, typically through verbalised confidence scores, token probabilities, or post-hoc calibration techniques \cite{Kale2025Mirage}. Despite recent progress, confidence signals remain imperfectly aligned with actual correctness, limiting their usefulness for downstream decision-making \cite{ren2026idontknowevaluating,kale2025knowrlteachinglanguagemodels}. More importantly, regardless of the estimation method, confidence is almost always treated as a property of a completed answer. We instead investigate whether confidence-related information encoded in hidden representations evolves throughout the answering process and can be anticipated before answer generation is complete.

\begin{figure*}[t]
        \centering
        \fbox{\includegraphics[width=0.65\linewidth]{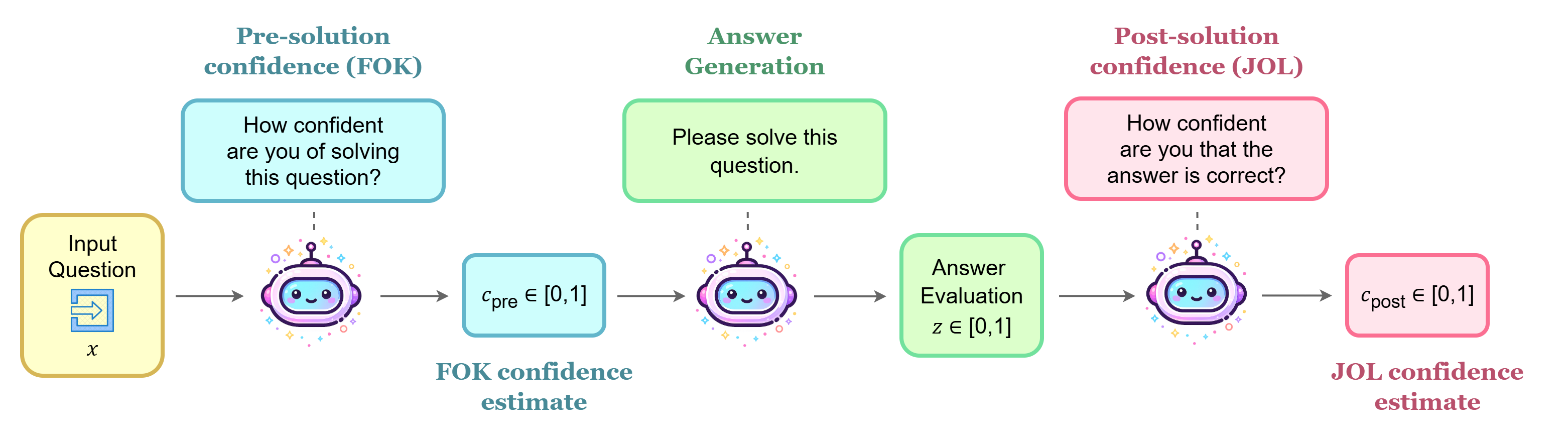}}
        \caption{Overview of temporal confidence measurement across answering stages}
        \label{fig:overview}
\end{figure*}

In this work, we study confidence from a temporal perspective. Inspired by classical theories of meta-memory \cite{nelson1990metamemory,koriat1997monitoring}, we distinguish between two stages of self-assessment: a pre-solution \emph{Feeling-of-Knowing} (FOK) signal and a post-solution confidence signal analogous to a \emph{Judgement-of-Learning} (JOL). The former reflects the model's estimate of future success before answering begins, while the latter reflects confidence after a solution has been produced. This framing allows us to analyse how confidence estimates and confidence-related information encoded in hidden representations change throughout the answering process. We first characterise the quality of both FOK and JOL signals across models and domains, and then investigate whether information present in post-solution hidden representations can be distilled into predictors operating on pre-solution hidden representations. Such an improved predictor would enable models to approximate the reliability of a completed answering process while operating solely on inexpensive pre-solution inputs, making it operationally convenient and much cheaper to deploy.

We hypothesise that post-solution hidden representations encode confidence-related information that is not fully recoverable from pre-solution confidence estimates alone. If so, post-solution probe confidence estimates could serve as supervisory signals for learning improved confidence predictors that operate entirely on pre-solution hidden representations. Specifically, we investigate whether future confidence predictors distilled from post-solution hidden representations can improve pre-solution confidence estimates, reduce calibration error, and generalise across common domains.

Across frontier and open-source LLMs spanning factual recall, logical reasoning, and mathematical reasoning, we compare verbal confidence estimates with representation-based confidence estimation. We show that post-solution confidence estimates are consistently better calibrated than pre-solution estimates, hidden representations encode substantially richer confidence-related information than models verbalise, and future confidence distillation recovers much of this improvement while operating entirely before answer generation. The main contributions of our work are as follows:
\begin{itemize}
\item We introduce a temporal perspective on confidence estimation in LLMs by distinguishing between pre-solution Feeling-of-Knowing (FOK) and post-solution Judgement-of-Learning (JOL) confidence signals, and provide a large-scale analysis of calibration, discrimination, and computational trade-offs across frontier and open-source models.

\item We introduce future confidence distillation, which trains predictors operating on pre-solution hidden representations using teacher confidence estimates from post-solution correctness probes, yielding substantially improved confidence estimation without requiring full answering generation.

\item We demonstrate that distilled future confidence predictors are sample efficient and transfer across unseen datasets within domains, suggesting that confidence-related information learned after answering can generalise beyond individual benchmarks.
\end{itemize}

\section{Related Work}

\subsection{Self-Knowledge and Confidence Estimation in LLMs}

Recent work has increasingly explored whether large language models exhibit forms of self-knowledge or metacognitive behaviour, referring to their ability to estimate the reliability of their own outputs and recognise uncertainty or knowledge limitations \cite{Kale2025Mirage,podolak-verma-2025-read,li-etal-2025-metacognition}. Existing studies show that LLMs can often predict whether they are likely to answer a question correctly, particularly when prompted to provide verbal confidence estimates \cite{kale2025knowrlteachinglanguagemodels}. However, these estimates frequently remain poorly calibrated, especially under reasoning-intensive tasks and distribution shift \cite{lin2022teaching,ren2026idontknowevaluating}. Reliable confidence estimation is increasingly important for downstream applications including retrieval augmentation, adaptive computation, tool use, model routing, and selective generation \cite{Wang2026SKR,zhang2026stopfailoperationalcapability}.

\subsection{Confidence Estimation from Hidden Representations}

Confidence estimation methods broadly fall into two categories. Black-box approaches infer confidence from behavioural signals such as verbal confidence estimates, self-evaluation prompts, answer consistency, and selective abstention \cite{xu-etal-2024-sayself,kale2025knowrlteachinglanguagemodels}. In contrast, white-box approaches exploit internal model signals including token probabilities, entropy, calibration statistics, and hidden representations \cite{Kadavath2022,malinin2021uncertaintyestimationautoregressivestructured,kissling2026selfawareknowledgeprobingevaluating}. Recent work has shown that linear probes trained on hidden representations often recover better calibrated confidence estimates than models explicitly verbalise \cite{kissling2026selfawareknowledgeprobingevaluating}, a theme we expand in our work.

\subsection{Temporal Confidence and Future Confidence Distillation}

Most prior work evaluates confidence after answer generation, treating confidence estimation as a property of completed responses. More recent studies show that pre-generation hidden representations already encode information about eventual answer correctness, with linear probes predicting future success before an answer is produced \cite{NoAnswerNeeded2025,LLMEncodeFailures2026}. Related work further demonstrates that correctness-related information remains linearly decodable throughout intermediate stages of answering, suggesting that outcome-relevant information emerges before the final prediction \cite{TemporalPredictors2025,LLMAlreadyKnows2025}.

While these studies focus on predicting correctness from hidden representations, they do not investigate how confidence-related information evolves throughout the answering process or an enhancement of this estimation. Our work addresses this gap by analysing temporal confidence representations and introducing future confidence distillation, which transfers confidence-related information from post-solution to pre-solution hidden representations.

\begin{figure*}[t]
    \begin{subfigure}{0.45\textwidth}
        \centering
        \includegraphics[width=\linewidth]{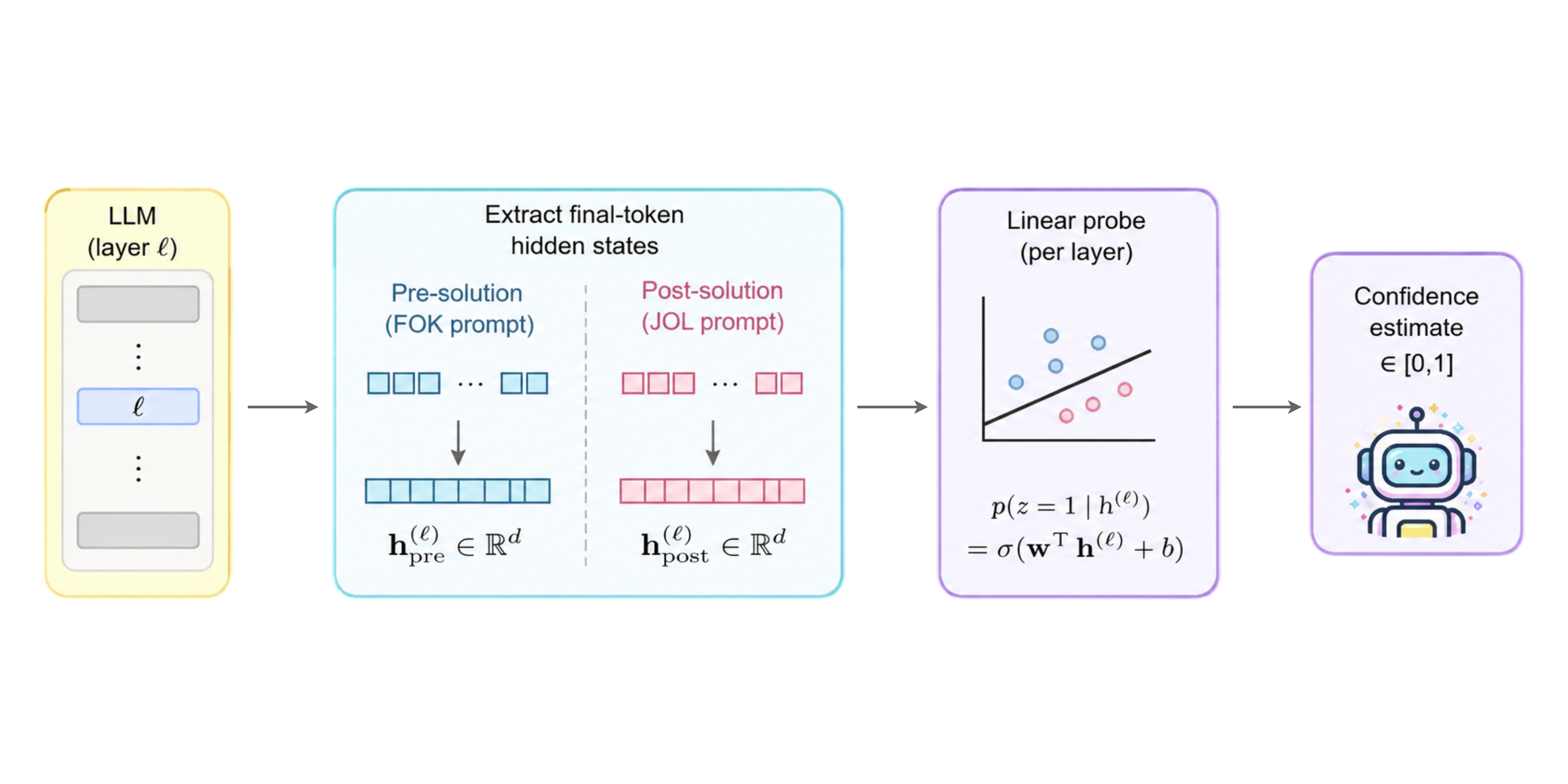}
        \caption{}
        \label{fig:probe}
    \end{subfigure}
    \hfill
    \begin{subfigure}{0.45\textwidth}
        \centering
        \includegraphics[width=\linewidth]{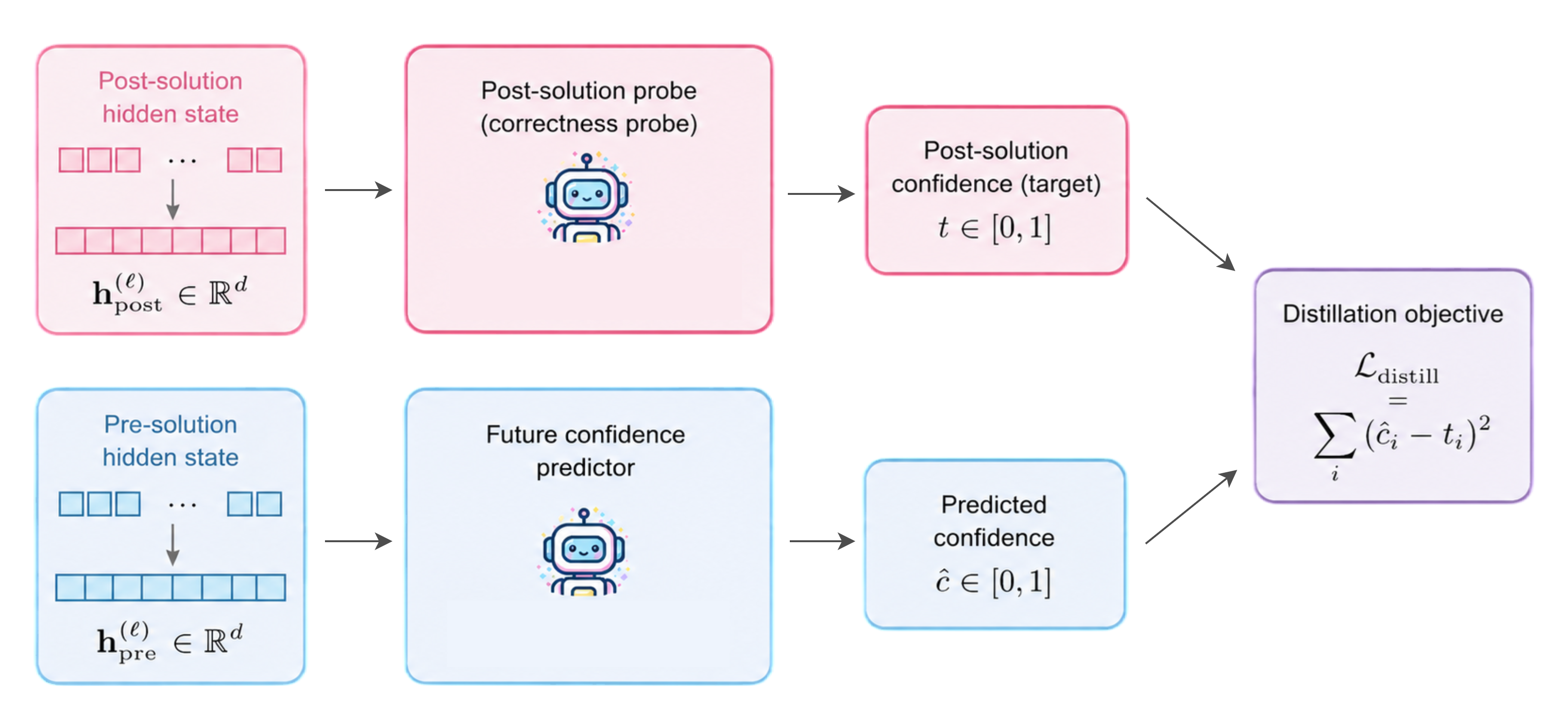}
        \caption{}
        \label{fig:future}
    \end{subfigure}

    \caption{Learning future confidence-related information encoded in hidden representations. (a) Hidden-state confidence probing. (b) Distillation from post-solution to pre-solution representations.}
    \label{fig:hero}
\end{figure*}

\section{Methodology}
\subsection{Temporal Confidence Signals}
Given an input problem $x$, we study confidence-related signals at two stages of the answering process, as shown in Fig~\ref{fig:overview}. 
\begin{itemize}
    \item Prior to answer generation, the model is asked to predict its probability of eventual success, yielding a pre-solution Feeling-of-Knowing (FOK) estimate $c_{\text{pre}}\in[0,1]$.
    \item The model then generates an answer $\hat{y}=f(x)$, which is evaluated against the reference solution to obtain a correctness label (based on domain) $z\in[0,1]$ or $z\in\{0,1\}$.
    \item Finally, the model is prompted to estimate the probability that its generated answer is correct, producing a post-solution Judgement-of-Learning (JOL) estimate $c_{\text{post}}\in[0,1]$.
\end{itemize}
Each example therefore yields $(x,\hat{y},z,c_{\text{pre}},c_{\text{post}})$, forming the basis of our calibration, representation, and distillation analyses. A summary of the notation used throughout the methodology is provided in Section \ref{app:notation} in the Appendix.

\subsection{Representation-Based Confidence Estimation}

We analyse confidence-related signals using both verbalised estimates and linear probes trained on hidden representations. Prompting yields verbal confidence estimates (FOK and JOL), providing a black-box confidence signal. Prompt templates are provided in Section \ref{app:prompts} in the Appendix.

Building on previous findings showing that linear probes trained on hidden representations often recover better calibrated confidence estimates from hidden representations \cite{kissling2026selfawareknowledgeprobingevaluating}, we extract the final-token hidden representations
$\mathbf{h}^{(\ell)}_{\text{pre}}, \mathbf{h}^{(\ell)}_{\text{post}}
\in \mathbb{R}^{d}$
from layer $\ell$ after processing the pre-solution and post-solution confidence prompts respectively, as shown in Fig~\ref{fig:probe}. For each layer, we train an independent linear probe

\[
p(z=1\mid\mathbf{h}^{(\ell)})=
\sigma(\mathbf{w}^{\top}\mathbf{h}^{(\ell)}+b),
\]

\noindent where $z$ denotes answer correctness and $\sigma(\cdot)$ is the logistic sigmoid. Probes are trained on disjoint train-test splits and evaluated using calibration and discrimination metrics. We primarily employ linear probes due to their strong empirical performance and interpretability, based on prior representation-probing work \cite{kissling2026selfawareknowledgeprobingevaluating,LLMAlreadyKnows2025}. We also evaluate non-linear probe architectures in Section \ref{app:non-linear} in the Appendix, but retain linear probes throughout because they consistently provide the strongest empirical performance while remaining readily interpretable.

\subsection{Future Confidence Distillation}

While post-solution confidence estimates consistently exhibit better calibration and discrimination (as analysed ahead), obtaining them requires generating a complete reasoning trajectory and answer. We therefore investigate whether predictors operating on pre-solution hidden representations can be trained to recover confidence-related information encoded in post-solution hidden representations to enable low-cost yet accurate confidence estimation, as displayed in Fig~\ref{fig:future}. 

For each layer $\ell$, we first train a post-solution probe

\[
t_i
=
T^{(\ell)}
\!\left(
\mathbf{h}^{(\ell)}_{\text{post},i}
\right)
\]

\noindent where $T^{(\ell)}$ is a correctness probe trained using labels $z$. The probe outputs a teacher confidence estimate, defined as the predicted probability of correctness. Unlike the binary correctness label, this continuous target captures confidence-related information encoded in post-solution hidden representations. We then train a distilled future confidence predictor

\[
\hat c_i
=
D^{(\ell)}
\!\left(
\mathbf{h}^{(\ell)}_{\text{pre},i}
\right)
\]

\noindent that operates exclusively on pre-solution representations and regresses toward post-solution confidence estimates. Specifically, we minimise

\[
\mathcal{L}_{\text{distill}}
=
\sum_i
\left(
\hat{c}_i
-
t_i
\right)^2.
\]

In practice, $T^{(\ell)}$ is implemented as a linear correctness probe and $D^{(\ell)}$ as a ridge regressor, a combination which performed most consistently across settings. Hidden representations are standardised and projected using PCA prior to training. Additional implementation details and architecture choices are provided in Section \ref{app:exp-impl} in the Appendix.

Unlike conventional knowledge distillation, which transfers task predictions between models, our approach transfers metacognitive information across answering stages within the same model. Specifically, the distilled predictor learns a mapping from pre-solution hidden representations to the teacher confidence estimates produced by a post-solution correctness probe. If successful, this allows models to approximate post-solution probe confidence estimates without generating a complete answering trajectory, substantially reducing inference cost while improving confidence calibration.

\section{Experimental Setup}
\subsection{Models}
We evaluate our approach on both frontier and open-source language models spanning multiple architectures and parameter scales. Frontier models are evaluated using verbal confidence estimates only, whereas open-source models additionally enable representation-based probing and future-confidence distillation through access to hidden representations.

For frontier black-box evaluation, we consider GPT-5.4 (\texttt{gpt-5.4-2026-03-05}) and Claude Sonnet 4.6 (\texttt{claude-sonnet-4-6}), accessed through the official OpenAI and Anthropic APIs respectively. These experiments enable analysis of confidence in proprietary systems where only verbal confidence estimates are available.

For representation and distillation experiments, we use five open-source instruction-tuned models: Llama-3.1-8B-Instruct, Qwen3-7B-Instruct, Mistral-Nemo-12B-Instruct, Qwen3-14B-Instruct, and Qwen3-32B-Instruct. Our selection is chosen to span multiple model families and scales ranging from 8B to 32B parameters, allowing analysis of how confidence-related information and distilled confidence predictors vary across models and scales. All experiments use deterministic decoding with temperature set to zero. Additional settings are in Section \ref{app:exp-impl} in the Appendix.

\subsection{Datasets}

We evaluate across nine datasets grouped into three domains: factual recall, logical reasoning, and mathematical reasoning (complete details in Section \ref{app:dataset} in the Appendix). 

\paragraph{Factual Recall.}
We use \texttt{TriviaQA} \cite{JoshiTriviaQA2017}, \texttt{Natural Questions} \cite{nqopen}, and \texttt{PopQA} \cite{popqa}. These datasets primarily evaluate parametric knowledge retrieval and factual recollection of trivia or entity-centric queries. 

\paragraph{Logical Reasoning.}
We use \texttt{RiddleBench} \cite{riddlebench}, \texttt{StrategyQA} \cite{stratgeyqa}, and \texttt{CommonsenseQA} \cite{commonsenseqa}. These benchmarks evaluate deductive reasoning, common-sense inference, and multi-step decision making. These tasks rely primarily on reasoning rather than factual recall, and therefore useful for studying confidence under demanding settings.

\paragraph{Mathematical Reasoning.}
We use \texttt{GSM8K} \cite{gsm8k}, \texttt{SVAMP} \cite{svamp}, and \texttt{MATH} \cite{math-dataset}. These datasets cover arithmetic word problems, compositional mathematical reasoning, and more challenging competition-style mathematics. Overall, these datasets provide a broad spectrum of mathematical difficulty, enabling analysis of how confidence estimation evolves as reasoning complexity increases.
\begin{figure*}[t]
    \centering
    \begin{subfigure}[t]{0.32\textwidth}
        \centering
        \includegraphics[width=\linewidth]{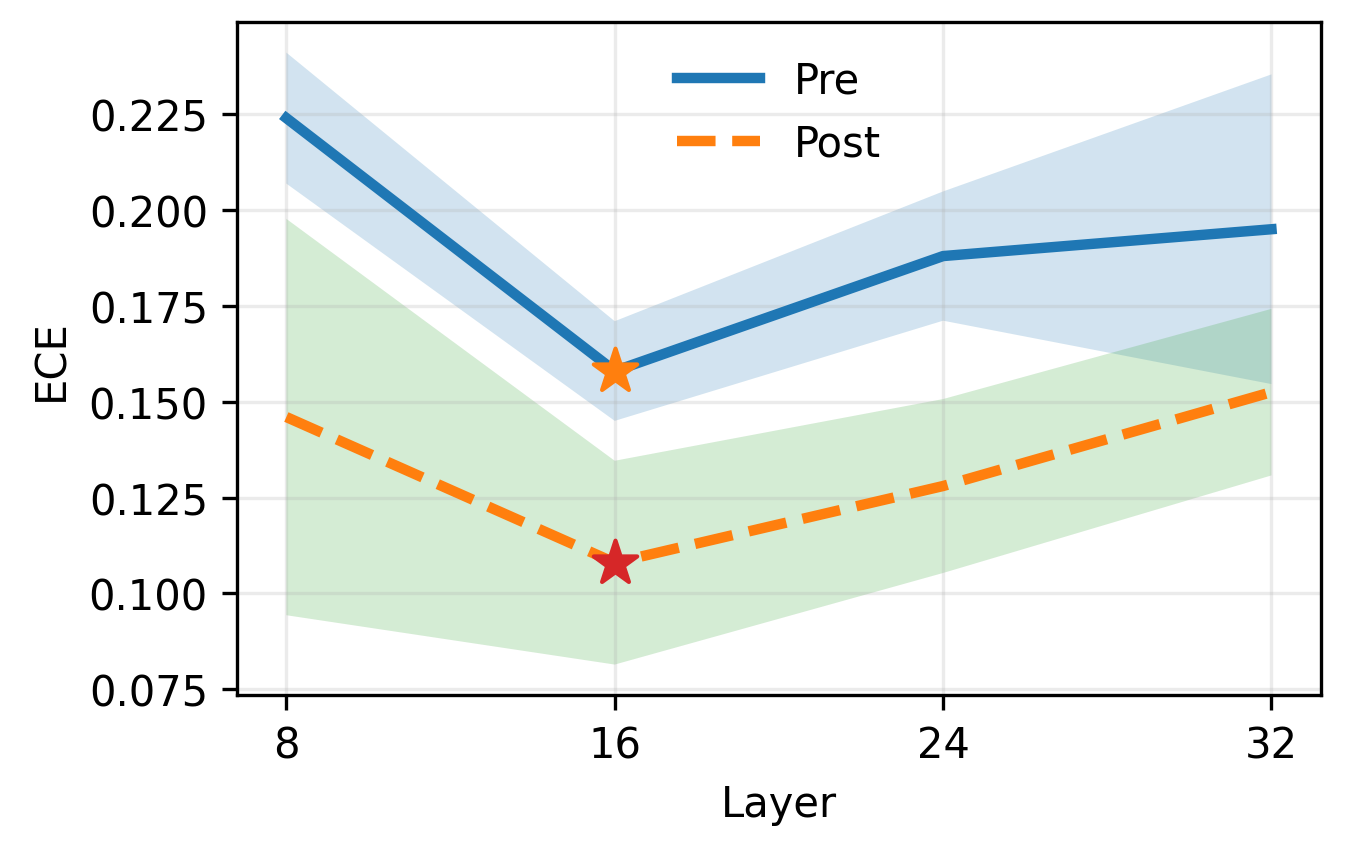}
        \caption{Factual recall}
    \end{subfigure}
    \hfill
    \begin{subfigure}[t]{0.32\textwidth}
        \centering
        \includegraphics[width=\linewidth]{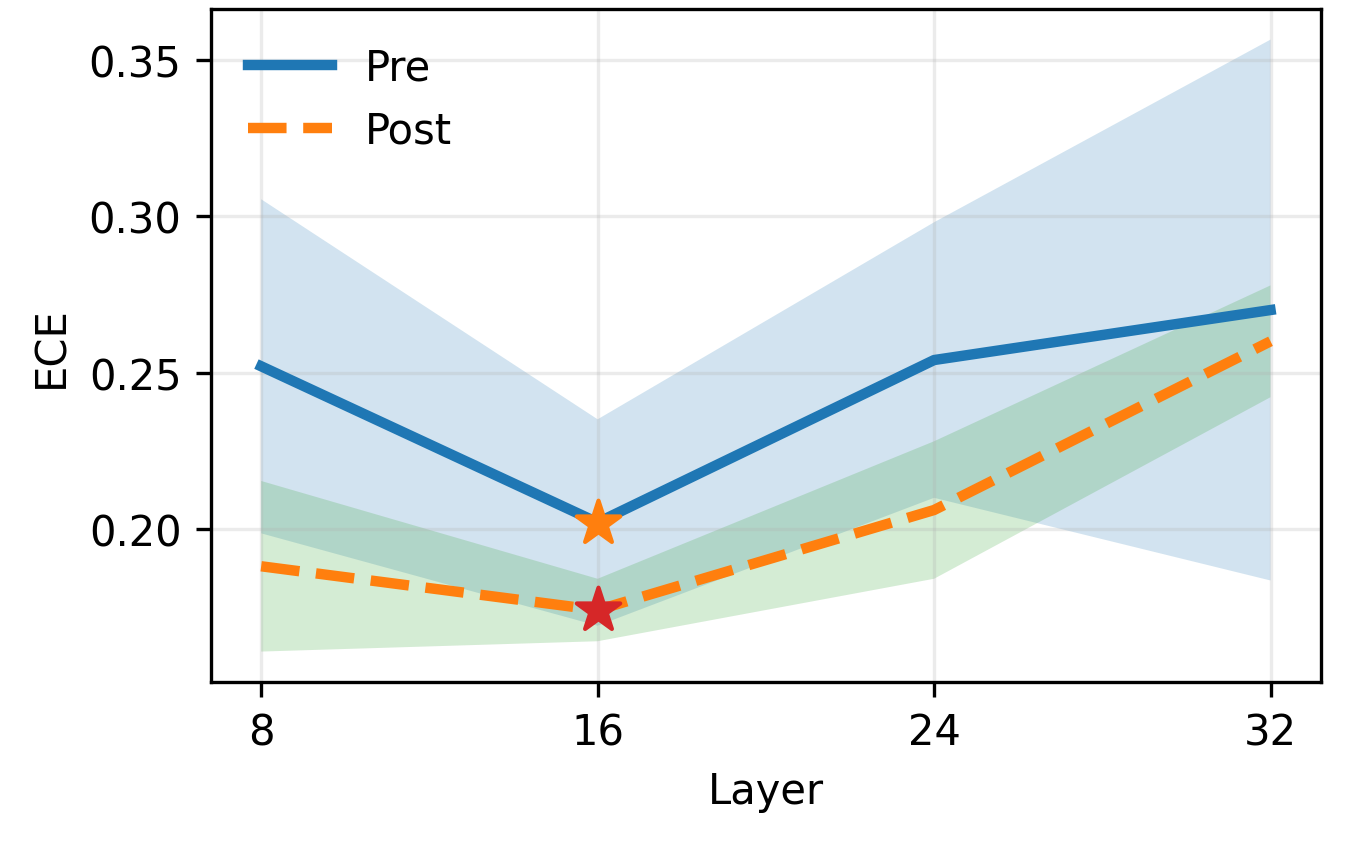}
        \caption{Logical reasoning}
    \end{subfigure}
    \hfill
    \begin{subfigure}[t]{0.32\textwidth}
        \centering
        \includegraphics[width=\linewidth]{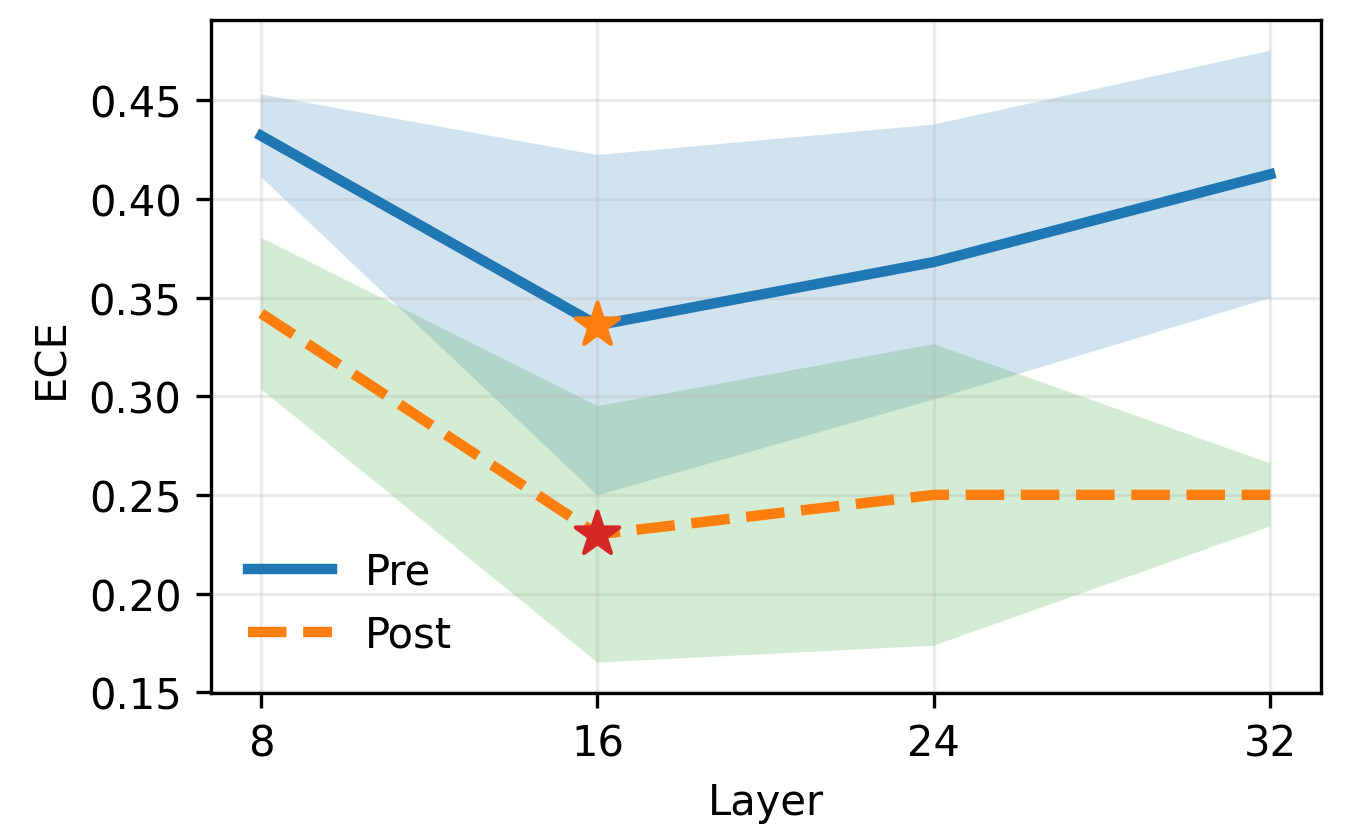}
        \caption{Mathematical reasoning}
    \end{subfigure}

    \caption{Layer-wise calibration (ECE) of linear probes trained on hidden representations averaged across open-source models. Solid lines denote pre-solution probes and dashed lines denote post-solution probes; shaded regions indicate 95\% confidence intervals. Stars mark the lowest mean ECE.}
    \label{fig:layerwise}
\end{figure*}

\subsection{Evaluation Protocol}

We evaluate verbal confidence estimates, probe confidence estimates, and distilled confidence estimates using both calibration and discrimination metrics, reporting Expected Calibration Error (ECE) and Area Under the Receiver Operating Characteristic Curve (AUROC). Unless otherwise stated, results are averaged across three random seeds and reported as mean scores with 95\% confidence intervals. To ensure stable calibration evaluation, we construct label-stratified dataset subsets such that no accuracy class exceeds 65\% of the examples, and preserve this distribution during probe train-test splitting. For linear probes trained on hidden representations, training is performed on disjoint train-test splits and evaluated independently at uniformly spaced network layers. Distilled future confidence predictors are further assessed under varying supervision budgets and transfer settings. Additional details are provided in Section \ref{app:metrics} in the Appendix.

\section{Results}
\subsection{How does confidence evolve during answering?}
\definecolor{accent}{RGB}{47,84,150}   
\definecolor{shade}{RGB}{244,246,250}  

\begin{table}[t]
\centering
\footnotesize
\setlength{\tabcolsep}{3pt}
\renewcommand{\arraystretch}{1.05}
\caption{Average verbal calibration (95\% CI). Lower ECE and higher AUROC indicate better calibration. Bold denotes the better value within each row.}
\label{tab:verbal_summary}

\begin{tabular}{@{}llcccc@{}}
\toprule
&&\multicolumn{2}{c}{\textbf{Pre-Solution}}&
\multicolumn{2}{c}{\textbf{Post-Solution}}\\
\cmidrule(lr){3-4}\cmidrule(lr){5-6}
\textbf{Models} & \textbf{Domain} &
ECE$\downarrow$ &
AUROC$\uparrow$ &
ECE$\downarrow$ &
AUROC$\uparrow$ \\
\midrule

\multirow{3}{*}{Closed}
& Factual & 0.14$\pm$0.05 & 0.61$\pm$0.06 & \textbf{0.13$\pm$0.04} & \textbf{0.66$\pm$0.05} \\
& Math    & 0.29$\pm$0.04 & 0.61$\pm$0.06 & \textbf{0.21$\pm$0.04} & \textbf{0.77$\pm$0.05} \\
& Logic   & 0.36$\pm$0.02 & \textbf{0.73$\pm$0.03} & \textbf{0.28$\pm$0.04} & 0.70$\pm$0.05 \\
\midrule

\multirow{3}{*}{Open}
& \cellcolor{shade}Factual & \cellcolor{shade}0.33$\pm$0.03 & \cellcolor{shade}0.61$\pm$0.04 &
\cellcolor{shade}\textbf{0.27$\pm$0.04} & \cellcolor{shade}\textbf{0.66$\pm$0.04} \\
& \cellcolor{shade}Math & \cellcolor{shade}0.33$\pm$0.04 & \cellcolor{shade}0.59$\pm$0.04 &
\cellcolor{shade}\textbf{0.28$\pm$0.05} & \cellcolor{shade}\textbf{0.62$\pm$0.03} \\
& \cellcolor{shade}Logic & \cellcolor{shade}0.38$\pm$0.03 & \cellcolor{shade}0.56$\pm$0.03 &
\cellcolor{shade}\textbf{0.35$\pm$0.03} & \cellcolor{shade}\textbf{0.59$\pm$0.04} \\
\bottomrule
\end{tabular}
\end{table}

Table \ref{tab:verbal_summary} compares verbal confidence estimates before (FOK) and after (JOL) answering across closed and open-source models for 1800 samples per domain, while Section \ref{app:verbal} in the Appendix details model-wise scores. A consistent pattern can be observed across nearly all settings, where verbal confidence estimated after answer generation is better calibrated and more predictive of correctness than confidence estimated beforehand, although the improvements are generally modest.

The largest improvements occur on mathematical tasks, where post-solution verbal confidence reduces calibration error while substantially increasing AUROC. This suggests that for domains requiring longer reasoning, the answering process exposes confidence-relevant information unavailable from the question representation alone. Factual recall exhibits smaller but consistent gains, reflecting the comparatively limited computation required once the relevant knowledge is already encoded. Despite these improvements, post-solution confidence estimates remain imperfectly calibrated across all domains, particularly for open-source models. This indicates that while the answer generation step improves expressed self-assessment, prompted confidence estimates may not still fully expose the metacognitive information available within the model.

\begin{table}[t]
\centering
\footnotesize
\setlength{\tabcolsep}{5pt}
\renewcommand{\arraystretch}{1.35}
\caption{Absolute performance gain of the strongest hidden-state probe confidence estimate relative to verbal confidence estimate on the same test set}
\label{tab:posttrain_gain}
\begin{tabular}{@{}l cc cc@{}}
\toprule
& \multicolumn{2}{c}{\textbf{Pre-Solution}} & \multicolumn{2}{c}{\textbf{Post-Solution}} \\
\cmidrule(lr){2-3}\cmidrule(lr){4-5}
\textbf{Domain} & \scriptsize $\Delta$ECE$\downarrow$ & \scriptsize $\Delta$AUROC$\uparrow$ &
\scriptsize $\Delta$ECE$\downarrow$ & \scriptsize $\Delta$AUROC$\uparrow$ \\
\midrule
Factual & \bst{$-$0.18} & \bmo{$+$0.11} & \bst{$-$0.17} & \bmo{$+$0.10} \\
Math    & \bnt{$-$0.01} & \bmo{$+$0.05} & \bmo{$-$0.06} & \bmo{$+$0.07} \\
Logic   & \bst{$-$0.20} & \bmo{$+$0.09} & \bst{$-$0.18} & \bst{$+$0.15} \\
\bottomrule
\end{tabular}
\end{table}

\subsection{Is metacognitive information hidden inside the representations beyond what the model verbalises?}
\label{sec:postvspre}
We next analyse whether the model's hidden representations encode richer confidence-related information than is expressed through verbal confidence estimates. Table \ref{tab:posttrain_gain} compares the best-performing layer-wise linear probe against verbal confidence estimates for open-source models on the held-out test set. For each model and dataset, probes are trained independently at eight-layer intervals using a fixed train-test split (1800 samples per domain, 60-40 split) and results are averaged over the best-performing layer (lowest validation ECE) for each model.

Across all three domains, layer-wise linear probes consistently outperform verbal confidence. This improvement is obtained without changing the underlying language model or prompting strategy, suggesting that confidence-related information is already encoded in hidden representations but is not fully expressed through verbal confidence. On factual recall and logical reasoning, calibration error is nearly halved while AUROC improves by 0.09-0.15. For mathematical reasoning, where calibration errors are largest, probes achieve modest yet consistent improvements in discrimination and calibration.

To understand where confidence is encoded, Figure~\ref{fig:layerwise} reports layer-wise probe calibration averaged across open-source models for each domain (full model-wise results in Section \ref{app:laywerise} in the Appendix). For every model, independent probes are trained at uniformly spaced eight-layer intervals using a fixed 60--40 train-test split (1800 examples per domain). Results show the mean test ECE with 95\% confidence intervals across models

Two consistent patterns emerge across all three domains. First, probes trained on post-solution hidden representations achieve lower calibration error than probes trained on pre-solution hidden representations throughout most of the network, indicating that confidence-related information encoded in hidden representations become progressively more decodable over the course of answering. Second, confidence-related information becomes most linearly decodable in intermediate layers, typically around layers 16--24, while calibration often deteriorates in the final decoding layers. This suggests that confidence-related information is encoded before the model commits to its final prediction, rather than arising solely from the output distribution.

\subsection{Can a low-cost future confidence estimator be distilled from pre-solution representations?}

\begin{table}[t]
\centering
\footnotesize
\setlength{\tabcolsep}{4pt}
\renewcommand{\arraystretch}{1.08}
\caption{Future confidence distillation averaged across five open-source LLMs. Recovery measures the fraction of the post--pre performance gap recovered by distillation.}
\label{tab:distill_main}
\begin{tabular}{lcccc}
\toprule
\textbf{AUROC}$\uparrow$ & Pre & Distilled & Post & Recovery \% \\
\midrule
Factual & 0.696 & 0.731 & 0.756 & 66.1\% \\
Logic   & 0.628 & 0.696 & 0.722 & 58.5\% \\
Math    & 0.634 & 0.652 & 0.668 & 31.7\% \\
\midrule
\textbf{ECE}$\downarrow$ & Pre & Distilled & Post & Recovery \% \\
\midrule
Factual & 0.161 & 0.133 & 0.104 & 54.9\% \\
Logic   & 0.202 & 0.184 & 0.170 & 38.0\% \\
Math    & 0.306 & 0.272 & 0.232 & 39.7\% \\
\bottomrule
\end{tabular}
\end{table}

\begin{figure*}[htb]
    \centering
    \begin{subfigure}[t]{0.32\textwidth}
        \centering
        \includegraphics[width=\linewidth]{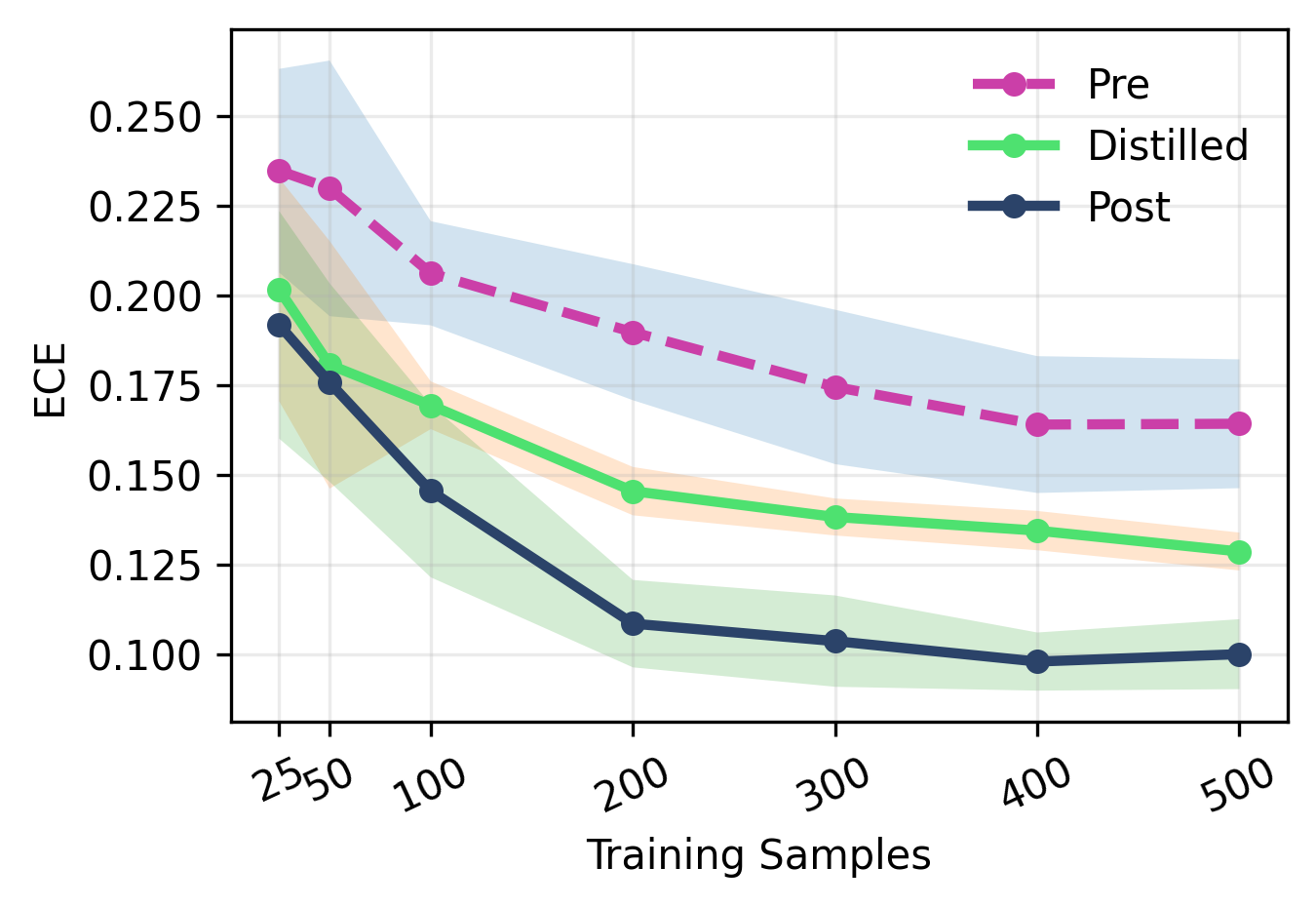}
        \caption{Factual recall}
    \end{subfigure}
    \hfill
    \begin{subfigure}[t]{0.32\textwidth}
        \centering
        \includegraphics[width=\linewidth]{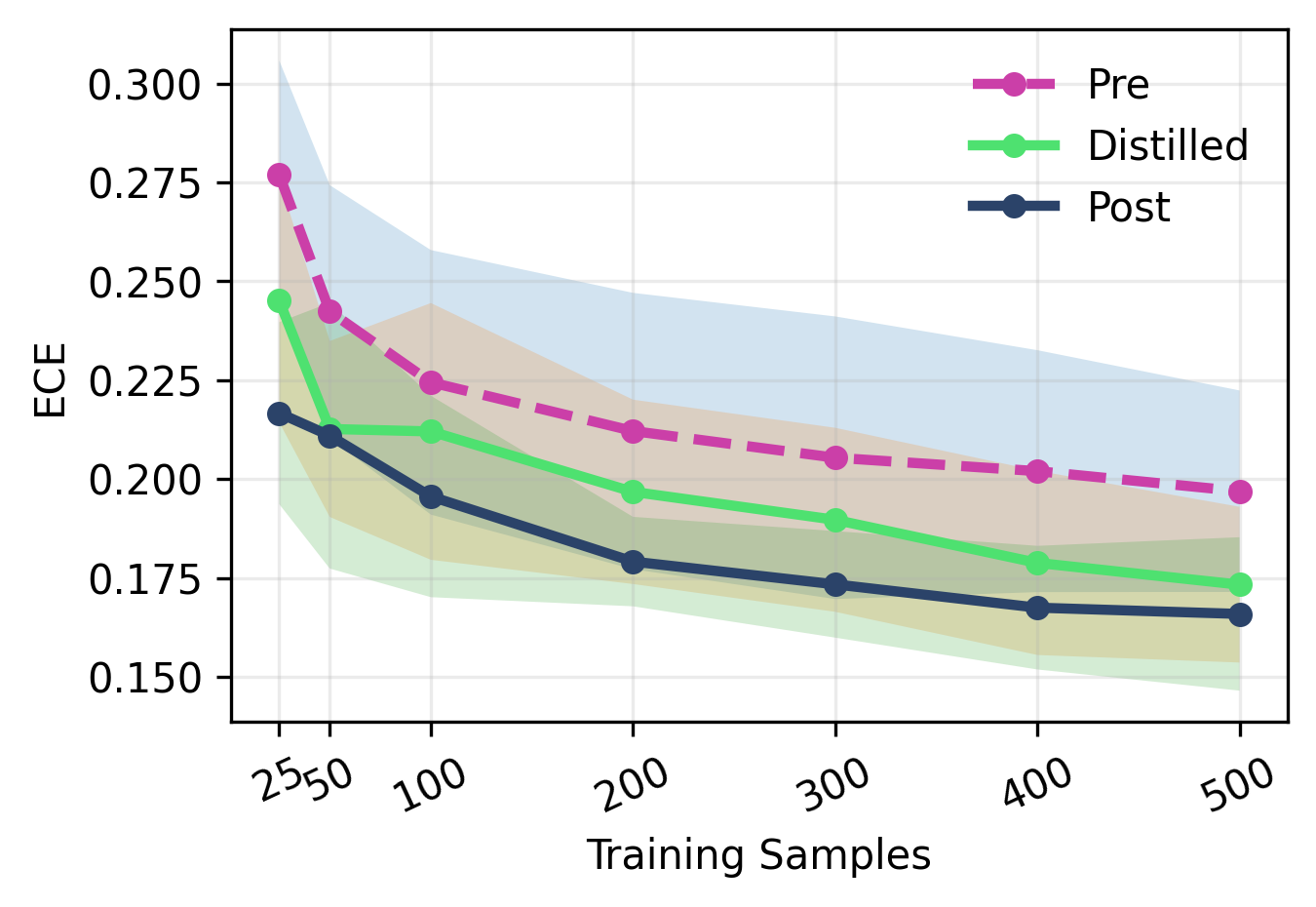}
        \caption{Logical reasoning}
    \end{subfigure}
    \hfill
    \begin{subfigure}[t]{0.32\textwidth}
        \centering
        \includegraphics[width=\linewidth]{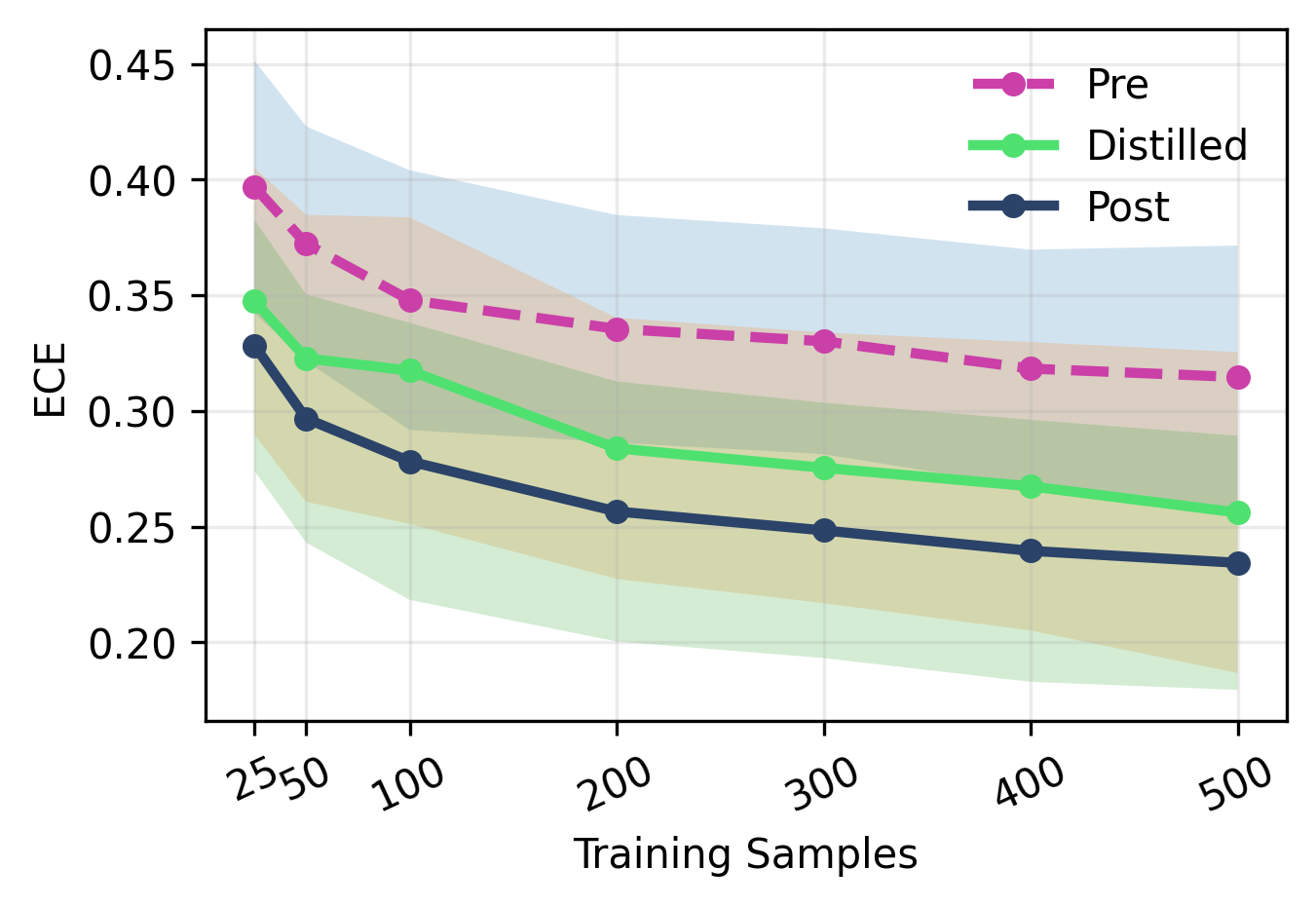}
        \caption{Mathematical reasoning}
    \end{subfigure}

    \caption{Calibration (ECE) as a function of training-set size. Pre- and post-solution probes are trained using correctness labels, whereas distilled predictors are trained using post-solution confidences. Curves show means across five open-source models and shaded regions denote 95\% confidence intervals.}
    \label{fig:sample_budget}
\end{figure*}

We next evaluate whether confidence estimates derived from post-solution hidden representations can be distilled into predictors operating on pre-solution hidden representations to obtain a computationally efficient and more calibrated confidence estimator. For each model and dataset, the post-solution probe is trained on hidden representations using a fixed 60--40 train-test split with 1800 examples. A ridge predictor is then trained to regress from pre-solution representations to the post-solution probe confidence estimates, with the operating layer selected using validation ECE.

Table~\ref{tab:distill_main} demonstrates that supervision from post-solution probes can be distilled into confidence predictors operating on pre-solution hidden representations quite effectively. Improvements are particularly pronounced on factual recall, despite exhibiting the smallest pre to post performance gap. Logical reasoning achieves moderate improvements, while mathematical reasoning remains the most challenging setting, but still exhibits consistent gains.

Our results suggest that probe confidence estimates serve as an effective supervisory signal rather than merely a stronger confidence estimate. Although the post-solution probe itself requires complete answer generation, much of the resulting confidence-related information can be predicted from pre-solution representations alone. Consequently, future confidence distillation provides substantially better calibrated confidence estimates with just pre-solution states as inputs to avoid computational costs during inference. 

\subsection{How sample efficient is confidence distillation?}
Since future confidence distillation offers a promising avenue of building low-cost and calibrated confidence predictors, we investigate the data efficiency of this process. For each model and domain, the post-solution probe is trained using the full training split, while distilled predictors are trained using progressively larger subsets of the available supervision (25-500 examples). Evaluation is performed on a fixed held-out test set of 500 examples from the same dataset, and results are averaged across five open-source models.

Figure~\ref{fig:sample_budget} plots ECE scores for probes trained with identical supervision budgets and reveals a consistent trend across all three domains (full model-wise results in Section \ref{app:same-dataset} in the Appendix). Even with only 25 labelled examples, distilled predictors already outperform directly supervised pre-solution probes with the same supervision budget. Performance improves steadily as additional supervision becomes available up to approximately 300-400 training examples, after which improvements largely plateau.

Factual recall is the most data-efficient setting, while logical reasoning follows a similar trajectory, achieving substantial improvements across the entire supervision range. Mathematical reasoning remains the most challenging domain, yet distillation consistently improves both AUROC and calibration even under extremely limited supervision. 

These results demonstrate that future confidence distillation is sample-efficient. Rather than requiring large labelled datasets, much of the confidence-related information encoded in post-solution hidden representations can be distilled into predictors operating on pre-solution hidden representations using only a few hundred training examples. This makes future confidence distillation practical in realistic settings where confidence supervision is limited.

\subsection{Can future confidence be learned once and transferred to unseen tasks?}

\begin{figure*}[htb]
    \centering
    \begin{subfigure}[t]{0.32\textwidth}
        \centering
        \includegraphics[width=\linewidth]{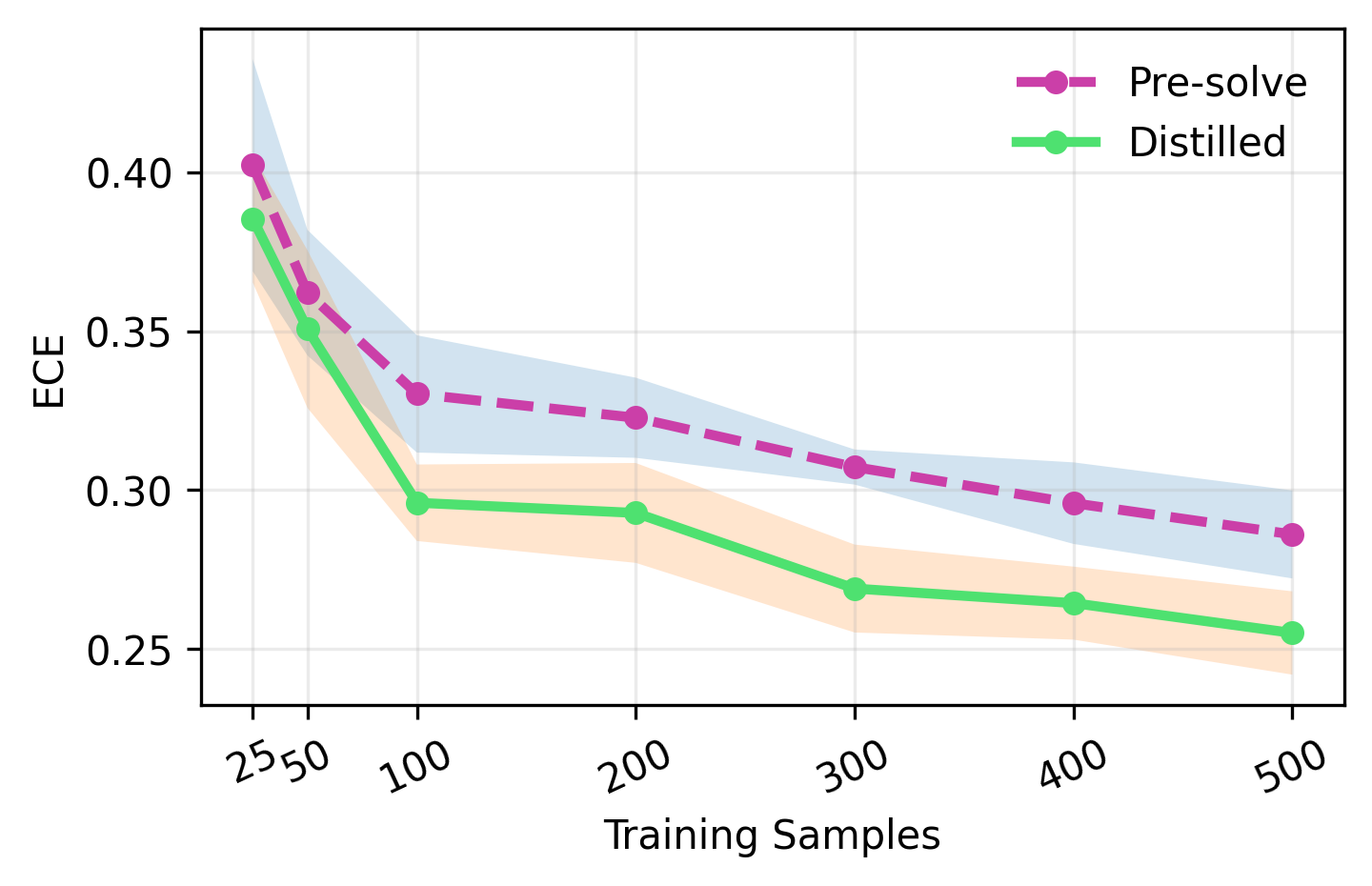}
        \caption{Factual recall}
    \end{subfigure}
    \hfill
    \begin{subfigure}[t]{0.32\textwidth}
        \centering
        \includegraphics[width=\linewidth]{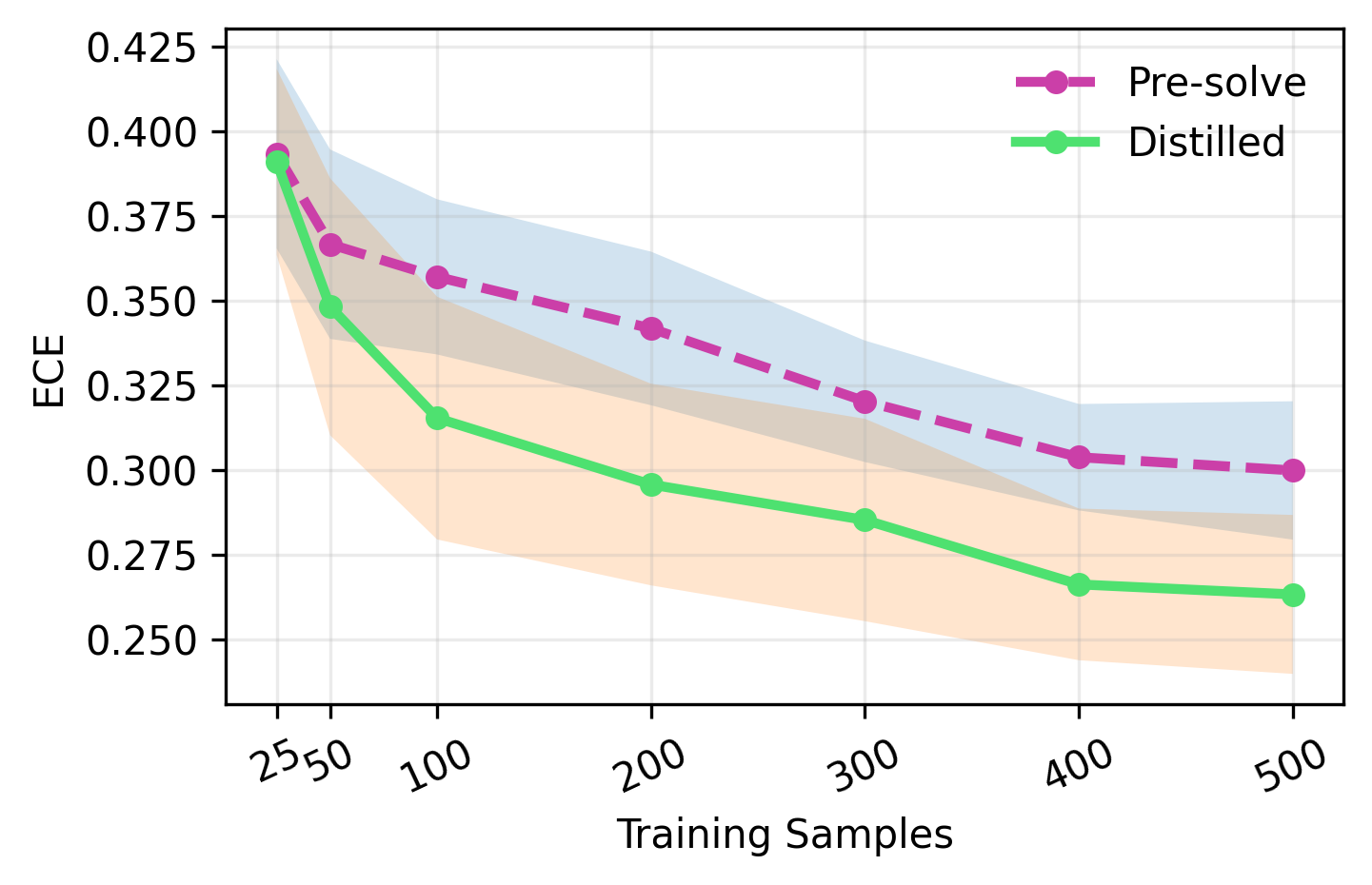}
        \caption{Logical reasoning}
    \end{subfigure}
    \hfill
    \begin{subfigure}[t]{0.32\textwidth}
        \centering
        \includegraphics[width=\linewidth]{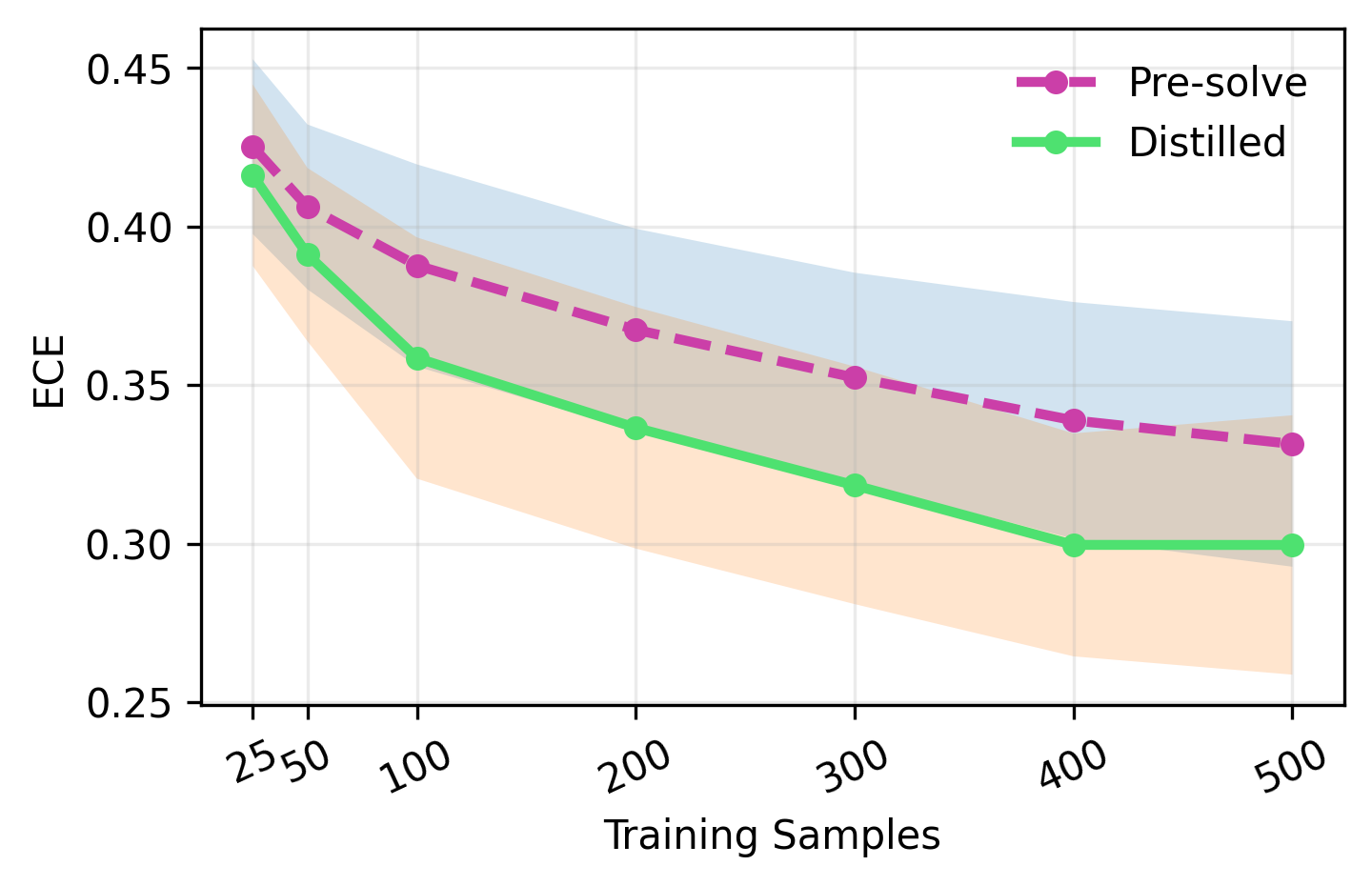}
        \caption{Mathematical reasoning}
    \end{subfigure}

    \caption{Calibration (ECE) as a function of training-set size for cross-dataset transfer within the same domain. Distilled predictors are trained on one dataset and evaluated on an unseen dataset without any target-domain fine-tuning. Curves show means across five open-source models and shaded regions denote 95\% confidence intervals.}
    \label{fig:cross_transfer}
\end{figure*}

Finally, we investigate whether distilled future confidence predictors generalise beyond a single benchmark. For each domain, distilled predictors are trained on one dataset using between 25 and 500 training examples and evaluated directly on a different dataset from the same domain without retraining. Results are averaged across five models and compared against pre-solution probes trained with the same supervision budget (full results in Section \ref{app:cross-dataset} in the Appendix).

Unlike the previous experiment, the predictor never observes examples from the target dataset during training and successful performance requires learning confidence predictors that generalise beyond dataset-specific statistics. Fig~\ref{fig:cross_transfer} shows that distilled future confidence predictors transfer reliably across unseen datasets within each domain. Although absolute performance is naturally lower than in the within-dataset setting, distilled predictors consistently achieve lower calibration error than directly supervised pre-solution probes. The improvement becomes increasingly pronounced as more supervision is available.

Transfer is strongest for factual recall, suggesting that confidence transfers more readily across benchmarks and has potential for direct improvement. Logical reasoning exhibits a similar trend despite substantial differences between reasoning tasks, and mathematical reasoning remains the most difficult, where improvements are smaller, yet mildly notable.

We also test cross-domain transfer (e.g., factual recall $\rightarrow$ mathematical reasoning), which performs substantially worse and is reported in Section \ref{app:cross-domain} in the Appendix, suggesting that distilled future confidence predictors generalise primarily within, rather than across domains.

Together, these results indicate that future confidence distillation learns reusable confidence predictors rather than dataset-specific decision boundaries. This suggests that future confidence can be learned once on representative tasks and transferred to new benchmarks within the same domain, substantially improving the practicality of calibrated confidence estimation at a much lower cost.

\section{Practical and Deployment Considerations}

\begin{table}[t]
\centering
\footnotesize
\caption{Relative deployment and inference requirements of the evaluated confidence estimators}
\label{tab:deployment}
\begin{tabular}{lccc}
\toprule
Estimator & Input & Rel. Inf. Cost \\
\midrule
Verbal FOK & Question  & 1$\times$\\
Pre Probe & Question  & 1$\times$\\
Future Confidence & Question  & 1$\times$\\
Verbal JOL & Question + Answer  & 2.3--2.5$\times$\\
Post Probe & Question + Answer & 2.2--2.5$\times$\\
\bottomrule
\end{tabular}
\end{table}

While post-solution confidence estimates consistently provide the strongest calibration, they require generating a complete answer before confidence can be assessed, increasing both inference latency and autoregressive decoding cost. Table~\ref{tab:deployment} compares the inference requirements of the evaluated confidence estimation methods. Verbal FOK, pre-solution probes, and the future confidence predictor operate directly on pre-solution hidden representations and therefore require only the input question at inference time. In contrast, verbal JOL and post-solution probes require complete answer generation and confidence estimation, increasing inference cost by approximately $2.3$--$2.5\times$ across domains.

Future confidence distillation provides a practical compromise between calibration and computational cost. Post-solution confidence estimates are used only offline to supervise training, while deployment requires only a lightweight predictor operating on pre-solution hidden representations. This preserves much of the calibration improvement obtained from post-solution confidence without incurring the additional cost of answer generation.

\section{Conclusion and Future Work}

We investigated how confidence evolves throughout the answering process and whether post-solution confidence can be anticipated before answering. Across both frontier and open-source LLMs, we found that post-solution verbal confidence estimates are consistently better calibrated than pre-solution estimates, while linear probes trained on hidden representations recover substantially richer confidence-related information than models can verbalise. Building on this observation, we introduced future confidence distillation, which learns a mapping from pre-solution hidden representations to post-solution probe confidence estimates. Despite operating entirely before answer generation at inference time, distilled predictors recover much of the calibration improvement achieved by post-solution confidence while remaining sample efficient and transferable. Together, these findings suggest that confidence-related information is recoverable and can be anticipated before the answering process is complete.

Future work includes extending distillation to larger reasoning models, multimodal systems, and agentic settings, as well as investigating stronger yet interpretable probe architectures and end-to-end training objectives. More broadly, our results highlight the potential of exploiting confidence-related information encoded in hidden representations to build more reliable and efficient language models.

\clearpage
\bibliography{aaai2027}

\appendix
\section{Notation Details}
\label{app:notation}
\begin{table}[b]
\centering
\small
\caption{Summary of notation used throughout the methodology.}
\label{tab:notation}
\begin{tabular}{ll}
\toprule
\textbf{Symbol} & \textbf{Description} \\
\midrule
$c_{\mathrm{pre}}$ & Pre-solution verbal confidence estimate (FOK) \\
$c_{\mathrm{post}}$ & Post-solution verbal confidence estimate (JOL) \\
$\mathbf{h}^{(\ell)}$ & Hidden representation at layer $\ell$ \\
$T^{(\ell)}$ & Post-solution correctness probe \\
$t_i$ & Teacher confidence estimate (output of $T^{(\ell)}$) \\
$D^{(\ell)}$ & Distilled confidence predictor \\
$\hat{c}_i$ & Distilled confidence estimate (output of $D^{(\ell)}$) \\
$z$ & Ground-truth correctness label \\
\bottomrule
\end{tabular}
\end{table}

We provide a short summary of the notations used throughout the methodology in Table~\ref{tab:notation}.

\section{Experimental Details}
\label{app:exp-impl}

\subsection{Dataset Details}
\label{app:dataset}
\begin{table*}[t]
\centering
\footnotesize
\setlength{\tabcolsep}{4pt}
\renewcommand{\arraystretch}{1.08}
\caption{Datasets used throughout the evaluation.}
\label{tab:appendix_datasets}
\begin{tabular}{llll}
\toprule
\textbf{Domain} & \textbf{Dataset} & \textbf{Task} & \textbf{Evaluation} \\
\midrule
\multirow{3}{*}{Factual}
& TriviaQA \cite{JoshiTriviaQA2017} & Open-domain QA & Exact Match \\
& Natural Questions \cite{nqopen} & Open-domain QA & Exact Match \\
& PopQA \cite{popqa} & Entity-centric QA & Exact Match \\
\midrule
\multirow{3}{*}{Logical}
& RiddleBench \cite{riddlebench} & Logical reasoning & Final answer match \\
& StrategyQA \cite{stratgeyqa} & Commonsense reasoning & Final answer match \\
& CommonsenseQA \cite{commonsenseqa} & Multiple-choice reasoning & Final answer match \\
\midrule
\multirow{3}{*}{Mathematical}
& GSM8K \cite{gsm8k} & Arithmetic reasoning & Exact Match \\
& SVAMP \cite{svamp} & Word problems & Exact Match \\
& MATH \cite{math-dataset} & Competition mathematics & Exact Match \\
\bottomrule
\end{tabular}
\end{table*}

We evaluate nine publicly available benchmarks spanning factual recall, logical reasoning, and mathematical reasoning, as shown in Table \ref{tab:appendix_datasets}. All datasets are obtained through the HuggingFace \texttt{datasets} library and used with a common prompting protocol across all models.

Unless otherwise stated, every experiment begins by randomly sampling 1800 examples from each dataset. Samples are partitioned into 60\% training and 40\% testing using three independent random seeds (0, 1, and 42), and all reported results are averaged across these splits. To select the best model layer whenever required (e.g., for probe selection or future confidence distillation), 10\% of the training partition is held out as a validation set, while all final metrics are reported exclusively on the unseen test partition.

Sample-efficiency experiments follow the same protocol but train using progressively smaller subsets (25--500 examples) drawn from the training partition while keeping the evaluation set fixed and separate. Cross-dataset transfer experiments train on samples mixed from 2 benchmarks and evaluate directly on the third different benchmark from the same reasoning domain without any target-domain fine-tuning.

Correctness labels are obtained using the standard evaluation protocol of each benchmark. Factual recall datasets are evaluated using normalised exact-match against the accepted reference answers. Logical reasoning datasets are evaluated by extracting the model's final predicted answer and comparing it with the ground-truth label. Mathematical reasoning benchmarks are evaluated using exact-match on the final numerical or symbolic answer. These correctness labels are used consistently throughout probe training, calibration evaluation, and future confidence distillation. To ensure fair comparison across domains and models, the same sampling protocol, prompt templates, train-validation-test partitioning strategy, and evaluation pipeline are used throughout all experiments unless explicitly stated otherwise.

\subsection{Language Models}
\label{app:models}
Table~\ref{tab:models_appendix} summarises all evaluated models. Frontier models are used only for behavioural confidence experiments because hidden representations are not publicly accessible. Representation probing and future confidence distillation are therefore conducted exclusively on open-source models.

\begin{table*}[h]
\centering
\footnotesize
\caption{Language models used throughout the paper.}
\label{tab:models_appendix}
\begin{tabular}{lll}
\toprule
\textbf{Model} & \textbf{Family} & \textbf{Use} \\
\midrule
GPT-5.4 \cite{singh2026openaigpt5card} & Closed-source & Verbal confidence only \\
Claude Sonnet 4.6 \cite{anthropic2026sonnet46} & Closed-source & Verbal confidence only \\
\midrule
Llama-3.1-8B-Instruct \cite{grattafiori2024llama3herdmodels} & Llama & Probing + Distillation \\
Qwen3-7B-Instruct \cite{qwen3} & Qwen & Probing + Distillation \\
Mistral-Nemo-12B-Instruct \cite{mistralai2024nemo} & Mistral & Probing + Distillation \\
Qwen3-14B-Instruct \cite{qwen3} & Qwen & Probing + Distillation \\
Qwen3-32B-Instruct \cite{qwen3} & Qwen & Probing + Distillation \\
\bottomrule
\end{tabular}
\end{table*}

\subsection{Inference Configuration}
\label{app:inf-config}
All experiments use deterministic decoding ($T=0$) to eliminate sampling variance during confidence estimation. Closed-source models are accessed through the official OpenAI and Anthropic APIs using their default decoding settings with temperature fixed to zero, and other parameters kept to default values.

Open-source models are loaded using the Unsloth framework with 4-bit quantisation for efficient inference with these settings:

\begin{itemize}
    \item 4-bit model loading
    \item maximum sequence length of 4096 tokens for answer generation
    \item maximum generation length of 32 tokens for confidence estimation
    \item maximum sequence length of 1024 tokens for confidence prompts
    \item greedy decoding ($\texttt{do\_sample=False}$)
\end{itemize}

Apart from these settings, all models use the default tokeniser and inference configuration supplied by their respective HuggingFace checkpoints.

\subsection{Hidden-State Extraction}
\label{app:hidden-state}
Hidden representations are extracted only from open-source models. For every confidence prompt, we perform a forward pass with \texttt{output\_hidden\_states=True} and extract the final-token representation from every sampled transformer layer before text generation.

To reduce storage and computation, hidden states are converted to \texttt{float16} after extraction. During probing experiments, representations are sampled at uniform eight-layer intervals (Layers 8, 16, 24, 32 where available), allowing comparable analysis across models with different depths.

All representation-based experiments use exactly the same hidden-state extraction procedure for both pre-solution and post-solution confidence prompts.

\subsection{Implementation Software and Setup}
\label{app:impl-software}
Future confidence predictors and confidence probes are implemented in Python using PyTorch, HuggingFace Transformers, Unsloth, scikit-learn, NumPy and the HuggingFace Datasets library.

All open-source experiments are performed two NVIDIA RTX 4090s (24+24 GB VRAM) running on Ubuntu 24.04 with CUDA 12.2. Hidden-state probes are trained using scikit-learn implementations of logistic regression and ridge regression after PCA projection and feature standardisation.

\subsection{Prompt Templates}
\label{app:prompts}
All confidence elicitation experiments use identical prompts across models and datasets. Only the user question changes between examples, while the prompt instructions remain fixed. For open-source models, prompts are formatted using each model's native chat template. Frontier models receive the equivalent system and user messages through their respective APIs.

\subsubsection{Pre-solution Confidence (FOK)}
\label{app:fok-prompt}
Before generating an answer, the model is instructed to estimate its probability of eventually solving the problem correctly without actually attempting the solution.

\begin{quote}
\small
\textbf{Prompt}

You are evaluating whether you can solve/answer a problem correctly. DO NOT solve the problem.

Briefly think about:
\begin{itemize}
\item reasoning complexity
\item number of steps required
\item possible failure modes
\end{itemize}

Output only: 

\texttt{Confidence: X\%}

where $X$ is a value between 0 and 100.
\end{quote}

\vspace{0.5em}

\subsubsection{Answer Generation}
\label{app:answer-prompt}
After recording the pre-solution confidence, the model receives the same problem again and is instructed to generate its final answer. A small additional instruction about answer formatting is placed depending on the domain.

\begin{quote}
\small
\textbf{Prompt}

Solve the problem carefully.

<If it is a mathematical problem, end your response with

\texttt{Final Answer: <answer>}>

<If it is a yes/no question or an MCQ, answer clearly with \texttt{Yes}, \texttt{No} or \texttt{<option>}.>
\end{quote}

\vspace{0.5em}

\subsubsection{Post-solution Confidence (JOL)}
\label{app:jol-prompt}
After answer generation, the model is prompted to estimate the probability that its own answer is correct.

\begin{quote}
\small
\textbf{System Prompt}

You already answered a given problem. Estimate the probability that your answer is correct.

Output only:

\texttt{Confidence: X\%}

where $X$ is a value between 0 and 100.
\end{quote}

Confidence values are extracted using a deterministic parser that identifies the first valid percentage in the generated response and normalises it to the interval $[0,1]$. Any additional generated text is ignored.

\section{Probe and Distillation Training Details}
\label{app:probe-train}
\subsection{Representation Preprocessing}
\label{app:rep-preprocess}
For every sampled transformer layer, the final-token hidden representation immediately preceding autoregressive generation is extracted and used as the feature vector. Following prior representation-probing work, hidden states are standardised using a \texttt{StandardScaler} fitted on the training split and subsequently projected to a lower-dimensional space using Principal Component Analysis (PCA). Unless otherwise stated, the PCA dimensionality is fixed to 128 components, with the projection fitted on the training data and applied unchanged to the validation and test sets.

To enable consistent comparison across models with different depths, representations are extracted at uniform eight-layer intervals throughout the network, always including the final transformer layer where applicable.

\subsection{Pre and Post-Solution Probe Training}
\label{app:pre-post-probe-train}
For every sampled layer, an independent linear correctness probe is trained to predict answer correctness from the corresponding hidden representation. We use logistic regression with the \texttt{lbfgs} solver, a maximum of 3000 optimisation iterations, and otherwise default scikit-learn hyperparameters. No architecture-specific hyperparameter tuning is performed.

Unless explicitly stated, for every dataset, 1800 randomly sampled examples are partitioned into a 60--40 train--test split. Ten percent of the training partition is further reserved as a validation set for model selection. The probe with the lowest validation Expected Calibration Error (ECE) is selected, and its performance is reported on the held-out test set. All experiments are repeated using three random seeds (0, 1, and 42), with reported results averaged across runs.

\subsection{Future Confidence Distillation}
\label{app:future-conf-train}
Future confidence predictors are trained independently at every sampled layer using the corresponding pre-solution hidden representations. The supervision target is the continuous confidence produced by the post-solution correctness probe trained on the same layer.

Distillation is implemented using ridge regression with the default scikit-learn regularisation settings. No hyperparameter tuning is performed beyond selecting the operating layer using validation ECE. As in the other experiments, training uses a 60--40 train--test split with 10\% of the training partition reserved for validation, and all reported results are averaged across three random seeds.

\subsection{Layer Selection}
\label{app:layer-select}
Rather than assuming that confidence is encoded uniformly throughout the network, probes and distilled predictors are trained independently at every sampled layer. The operating layer is selected using validation ECE, after which the corresponding model is evaluated once on the held-out test split.

This procedure is repeated independently for every model, dataset, and random seed, allowing the optimal confidence representation to vary across architectures and reasoning domains.

\subsection{Evaluation Metrics}
\label{app:metrics}
Probe performance is evaluated using both Expected Calibration Error (ECE) and Area Under the Receiver Operating Characteristic Curve (AUROC). ECE is computed using ten equally spaced confidence bins, while AUROC is calculated using the \texttt{roc\_auc\_score} implementation provided by scikit-learn.

Unless otherwise stated, reported values correspond to the mean across three random seeds, with 95\% confidence intervals computed over the corresponding experimental repetitions. Aggregated figures further average results across datasets or models, depending on the experiment described in the main text.

\section{Probe Architecture Ablation}
\label{app:non-linear}

\begin{table}[t]
\centering
\footnotesize
\caption{Probe architecture comparison. Values denote mean ECE averaged across five open-source models (600 training / 400 test examples per domain). Lower is better. Best architecture for each task is highlighted.}
\label{tab:probe_ablation}
\begin{tabular}{llccc}
\toprule
Task & Model & Factual & Logic & Math\\
\midrule
\multirow{4}{*}{Pre Probe}
& \textbf{Logistic} & \textbf{0.156} & \textbf{0.231} & \textbf{0.275}\\
& Ridge & 0.220 & 0.308 & 0.331\\
& MLP (1 layer) & 0.354 & 0.442 & 0.461\\
& MLP (2 layer) & 0.401 & 0.408 & 0.456\\
\midrule
\multirow{4}{*}{Post Probe}
& \textbf{Logistic} & \textbf{0.104} & \textbf{0.153} & \textbf{0.224}\\
& Ridge & 0.222 & 0.263 & 0.332\\
& MLP (1 layer) & 0.331 & 0.451 & 0.481\\
& MLP (2 layer) & 0.392 & 0.401 & 0.478\\
\midrule
\multirow{4}{*}{Future Confidence}
& Logistic & 0.183 & 0.228 & 0.297\\
& \textbf{Ridge} & \textbf{0.121} & \textbf{0.183} & \textbf{0.250}\\
& MLP (1 layer) & 0.337 & 0.397 & 0.492\\
& MLP (2 layer) & 0.386 & 0.442 & 0.542\\
\bottomrule
\end{tabular}
\end{table}

To justify the choice of probe architecture, we compare several commonly used linear and non-linear predictors under identical training conditions. Experiments use 600 training and 400 held-out test examples per domain, averaged across the five evaluated open-source models.

Table~\ref{tab:probe_ablation} shows that linear models consistently achieve the best calibration. Logistic regression performs best for both pre-solution and post-solution correctness probing, while ridge regression is most effective for future confidence distillation. In contrast, both one- and two-layer multilayer perceptrons substantially increase calibration error across every domain.

The consistent advantage of linear models suggests that confidence-related information encoded in hidden representations is already largely linearly separable within hidden states. Increasing model capacity primarily introduces overfitting without improving calibration, particularly under the relatively small supervision budgets considered in this work. Consequently, all experiments in the main paper employ logistic regression for correctness probes and ridge regression for future confidence distillation.

\section{Additional Experimental Results}
\label{app:add-exps}

\subsection{Model-wise Verbal Confidence Performance}
\label{app:verbal}

Tables \ref{tab:closed_model_results} and \ref{tab:open_model_results} present the comprehensive verbal confidence results for all evaluated models across every reasoning domain. These results are averaged over three random seeds and are reported as the mean $\pm$ standard deviation.The trends observed in the main paper remain consistent across individual models. 

Key takeaways include:
\begin{itemize}
    \item Calibration and Discrimination: Post-solution confidence generally achieves better calibration and discrimination than pre-solution confidence, though the magnitude of this improvement varies by model family and reasoning domain
    \item Domain-Specific Performance: Factual recall typically exhibits the smallest calibration gap between pre- and post-solution signals. In contrast, logical and mathematical reasoning benefit more substantially from post-solution self-assessment
\end{itemize}

\begin{table*}[t]
\centering
\footnotesize
\caption{Model-wise verbal confidence performance for frontier closed-source models. Lower ECE is better and higher AUROC is better. Results are mean $\pm$ standard deviation across three random seeds.}
\label{tab:closed_model_results}

\setlength{\tabcolsep}{5pt}
\begin{tabular}{llcccc}
\toprule
Model & Domain &
Pre ECE$\downarrow$ &
Post ECE$\downarrow$ &
Pre AUROC$\uparrow$ &
Post AUROC$\uparrow$\\
\midrule

GPT-5.4
& Factual & $0.20\pm0.03$ & $0.19\pm0.02$ & $0.57\pm0.05$ & $0.61\pm0.05$\\
& Logic   & $0.38\pm0.04$ & $0.29\pm0.04$ & $0.68\pm0.06$ & $0.69\pm0.05$\\
& Math    & $0.38\pm0.04$ & $0.32\pm0.05$ & $0.72\pm0.05$ & $0.75\pm0.05$\\
\midrule

Claude Sonnet 4.6
& Factual & $0.07\pm0.02$ & $0.08\pm0.02$ & $0.66\pm0.06$ & $0.70\pm0.05$\\
& Logic   & $0.34\pm0.05$ & $0.27\pm0.05$ & $0.77\pm0.06$ & $0.70\pm0.05$\\
& Math    & $0.19\pm0.06$ & $0.11\pm0.06$ & $0.50\pm0.09$ & $0.79\pm0.09$\\

\bottomrule
\end{tabular}
\end{table*}

\begin{table*}[t]
\centering
\footnotesize
\caption{Model-wise verbal confidence performance for open-source instruction-tuned models. Lower ECE is better and higher AUROC is better. Results are mean $\pm$ standard deviation across three random seeds.}
\label{tab:open_model_results}

\setlength{\tabcolsep}{4pt}
\begin{tabular}{llcccc}
\toprule
Model & Domain &
Pre ECE$\downarrow$ &
Post ECE$\downarrow$ &
Pre AUROC$\uparrow$ &
Post AUROC$\uparrow$\\
\midrule

Llama-3.1-8B
& Factual & 0.29$\pm$0.02 & 0.23$\pm$0.06 & 0.59$\pm$0.04 & 0.67$\pm$0.05\\
& Logic   & 0.41$\pm$0.06 & 0.39$\pm$0.05 & 0.52$\pm$0.07 & 0.55$\pm$0.06\\
& Math    & 0.34$\pm$0.05 & 0.31$\pm$0.04 & 0.53$\pm$0.07 & 0.57$\pm$0.04\\
\midrule

Qwen3-7B
& Factual & 0.32$\pm$0.03 & 0.32$\pm$0.04 & 0.56$\pm$0.07 & 0.60$\pm$0.04\\
& Logic   & 0.38$\pm$0.06 & 0.35$\pm$0.04 & 0.57$\pm$0.05 & 0.59$\pm$0.03\\
& Math    & 0.41$\pm$0.04 & 0.37$\pm$0.06 & 0.51$\pm$0.05 & 0.53$\pm$0.05\\
\midrule

Mistral-Nemo-12B
& Factual & 0.47$\pm$0.04 & 0.31$\pm$0.03 & 0.60$\pm$0.05 & 0.64$\pm$0.06\\
& Logic   & 0.37$\pm$0.07 & 0.33$\pm$0.06 & 0.53$\pm$0.07 & 0.58$\pm$0.07\\
& Math    & 0.27$\pm$0.05 & 0.22$\pm$0.04 & 0.62$\pm$0.06 & 0.65$\pm$0.05\\
\midrule

Qwen3-14B
& Factual & 0.32$\pm$0.05 & 0.29$\pm$0.04 & 0.64$\pm$0.05 & 0.70$\pm$0.05\\
& Logic   & 0.39$\pm$0.05 & 0.35$\pm$0.05 & 0.56$\pm$0.06 & 0.60$\pm$0.05\\
& Math    & 0.32$\pm$0.06 & 0.27$\pm$0.04 & 0.63$\pm$0.03 & 0.65$\pm$0.05\\
\midrule

Qwen3-32B
& Factual & 0.25$\pm$0.04 & 0.22$\pm$0.04 & 0.67$\pm$0.04 & 0.71$\pm$0.06\\
& Logic   & 0.35$\pm$0.05 & 0.33$\pm$0.06 & 0.59$\pm$0.06 & 0.63$\pm$0.06\\
& Math    & 0.31$\pm$0.06 & 0.27$\pm$0.05 & 0.66$\pm$0.05 & 0.69$\pm$0.05\\

\bottomrule
\end{tabular}
\end{table*}

\subsection{Layer-wise Probe Performance}
\label{app:laywerise}

Tables \ref{tab:layer_factual} through \ref{tab:layer_math} report probe performance at each evaluated layer for every open-source model, utilising the same probe training protocol described before.

The model-level results closely mirror the aggregate trends presented in the main paper. Across nearly all models, post-solution representations consistently achieve lower calibration error than pre-solution representations at corresponding layers. The strongest performance emerges in intermediate representations, typically between layers 16 and 24, while calibration performance often deteriorates toward the final decoding layers. Overall, these trends remain consistent across various model families and parameter scales, suggesting that the location of metacognitive information may be largely architecture-independent.

\begin{table*}[t]
\centering
\footnotesize
\caption{Layer-wise probe performance on factual recall. Lower ECE is better and higher AUROC is better.}
\label{tab:layer_factual}

\setlength{\tabcolsep}{6pt}
\begin{tabular}{lccccc}
\toprule
Model & Layer & Pre ECE$\downarrow$ & Post ECE$\downarrow$ & Pre AUROC$\uparrow$ & Post AUROC$\uparrow$\\
\midrule

Llama-3.1-8B
& 8  &0.21&0.14&0.72&0.76\\
&16 &\textbf{0.16}&\textbf{0.09}&0.73&0.79\\
&24 &0.18&0.10&0.74&0.73\\
&32 &0.19&0.18&0.68&0.66\\
\midrule

Qwen3-7B
&8 &0.21&0.19&0.72&0.70\\
&16&0.15&0.15&0.74&0.71\\
&24&0.19&\textbf{0.12}&0.74&0.77\\
\midrule

Mistral-Nemo-12B
&8 &0.21&0.10&0.66&0.71\\
&16&\textbf{0.18}&\textbf{0.08}&0.68&0.77\\
&24&0.22&0.11&0.70&0.75\\
&32&0.25&0.13&0.69&0.77\\
\midrule

Qwen3-14B
&8 &0.24&\textbf{0.08}&0.33&0.69\\
&16&\textbf{0.14}&0.09&0.70&0.79\\
&24&0.18&0.16&0.73&0.75\\
&32&0.19&0.16&0.74&0.78\\
\midrule

Qwen3-32B
&8 &0.25&0.22&0.67&0.68\\
&16&0.16&\textbf{0.13}&0.70&0.78\\
&24&0.17&0.15&0.69&0.72\\
&32&\textbf{0.15}&0.14&0.74&0.76\\

\bottomrule
\end{tabular}
\end{table*}

\begin{table*}[t]
\centering
\footnotesize
\caption{Layer-wise probe performance on logical reasoning. Lower ECE is better and higher AUROC is better.}
\label{tab:layer_logical}

\setlength{\tabcolsep}{5pt}
\begin{tabular}{llcccc}
\toprule
Model & Layer & Pre ECE$\downarrow$ & Post ECE$\downarrow$ & Pre AUROC$\uparrow$ & Post AUROC$\uparrow$\\
\midrule

Llama-3.1-8B
& 8  & 0.29 & 0.24 & 0.52 & 0.66\\
& 16 & 0.24 & \textbf{0.19} & 0.63 & 0.68\\
& 24 & 0.33 & 0.20 & 0.54 & 0.63\\
& 32 & 0.39 & 0.28 & 0.52 & 0.65\\
\midrule

Qwen3-7B
& 8  & 0.29 & \textbf{0.16} & 0.52 & 0.70\\
& 16 & 0.20 & 0.17 & 0.60 & 0.64\\
& 24 & 0.25 & 0.23 & 0.54 & 0.66\\
\midrule

Mistral-Nemo-12B
& 8  & 0.24 & 0.19 & 0.64 & 0.68\\
& 16 & \textbf{0.16} & 0.18 & 0.68 & 0.69\\
& 24 & 0.24 & 0.17 & 0.64 & 0.73\\
& 32 & 0.19 & 0.27 & 0.64 & 0.67\\
\midrule

Qwen3-14B
& 8  & 0.29 & \textbf{0.17} & 0.62 & 0.74\\
& 16 & 0.17 & \textbf{0.17} & 0.66 & 0.78\\
& 24 & 0.19 & 0.23 & 0.64 & 0.76\\
& 32 & 0.29 & 0.24 & 0.63 & 0.80\\
\midrule

Qwen3-32B
& 8  & \textbf{0.15} & 0.18 & 0.67 & 0.80\\
& 16 & 0.24 & \textbf{0.16} & 0.65 & 0.79\\
& 24 & 0.26 & 0.20 & 0.62 & 0.75\\
& 32 & 0.22 & 0.25 & 0.60 & 0.74\\

\bottomrule
\end{tabular}
\end{table*}

\begin{table*}[t]
\centering
\footnotesize
\caption{Layer-wise probe performance on mathematical reasoning. Lower ECE is better and higher AUROC is better.}
\label{tab:layer_math}

\setlength{\tabcolsep}{6pt}
\begin{tabular}{lccccc}
\toprule
Model & Layer & Pre ECE$\downarrow$ & Post ECE$\downarrow$ & Pre AUROC$\uparrow$ & Post AUROC$\uparrow$\\
\midrule

Llama-3.1-8B
&8  &0.39&0.27&0.56&0.60\\
&16 &\textbf{0.33}&\textbf{0.22}&0.60&0.64\\
&24 &0.38&0.23&0.55&0.62\\
&32 &0.44&0.25&0.56&0.57\\
\midrule

Qwen3-7B
&8  &0.44&0.39&0.57&0.56\\
&16 &0.50&\textbf{0.36}&0.55&0.58\\
&24 &0.47&0.40&0.54&0.55\\
\midrule

Mistral-Nemo-12B
&8  &0.44&0.35&0.61&0.59\\
&16 &\textbf{0.33}&\textbf{0.18}&0.68&0.71\\
&24 &0.39&0.24&0.63&0.65\\
&32 &0.49&0.27&0.55&0.67\\
\midrule

Qwen3-14B
&8  &0.45&0.35&0.60&0.58\\
&16 &\textbf{0.25}&0.20&0.62&0.66\\
&24 &\textbf{0.25}&\textbf{0.18}&0.65&0.67\\
&32 &0.36&0.23&0.59&0.61\\
\midrule

Qwen3-32B
&8  &0.44&0.35&0.60&0.69\\
&16 &\textbf{0.27}&\textbf{0.19}&0.67&0.74\\
&24 &0.35&0.20&0.64&0.75\\
&32 &0.36&0.25&0.65&0.73\\

\bottomrule
\end{tabular}
\end{table*}

\subsection{Sample Efficiency Across Model Families}
\label{app:same-dataset}

To examine whether future confidence distillation is consistently sample-efficient across model families, we repeat the training-budget experiments for each individual open-source model. Results are averaged across the three reasoning domains using training sets ranging from 25 to 500 examples. Table~\ref{tab:model_budget} reports the lowest and highest supervision budgets, while the complete learning curves are presented in the main paper.

Across all five models, distilled predictors consistently outperform directly supervised pre-solution probes under both extremely limited and abundant supervision. Larger models generally achieve lower calibration error overall, but every model benefits from future confidence supervision. As training data increases, distilled predictors progressively approach the calibration achieved by post-solution probes while retaining the deployment cost of pre-solution estimation.

\begin{table*}[t]
\centering
\footnotesize
\caption{Calibration (ECE) across model families under low-resource (25 examples) and high-resource (500 examples) supervision. Lower is better. Results are averaged across all three reasoning domains.}
\label{tab:model_budget}

\setlength{\tabcolsep}{6pt}
\begin{tabular}{lcccccc}
\toprule
& \multicolumn{3}{c}{25 Training Examples} &
\multicolumn{3}{c}{500 Training Examples}\\
\cmidrule(lr){2-4}\cmidrule(lr){5-7}

Model
& Pre
& Distilled
& Post
& Pre
& Distilled
& Post\\

\midrule

Llama-3.1-8B
&0.337&0.276&0.256
&0.231&0.192&0.171\\

Qwen3-7B
&0.319&0.295&0.286
&0.261&0.229&0.203\\

Mistral-Nemo-12B
&0.311&0.270&0.244
&0.246&0.175&0.161\\

Qwen3-14B
&0.268&0.244&0.211
&0.200&0.170&0.154\\

Qwen3-32B
&0.280&0.238&0.230
&0.189&0.164&0.144\\

\bottomrule
\end{tabular}
\end{table*}

\subsection{Cross-Dataset Transfer Across Model Families}
\label{app:cross-dataset}

We further examine whether future confidence-related information encoded in hidden representations transfers consistently across model families when trained and evaluated on different datasets within the same reasoning domain. For each model, the future confidence predictor is trained on one dataset and evaluated without retraining on a disjoint dataset from the same domain. Results are averaged across the three domains.

Table~\ref{tab:model_transfer} reports the lowest and highest supervision budgets. Unlike the within-dataset setting, a small number of models exhibit marginal degradation under only 25 training examples. However, these differences disappear rapidly with additional supervision, after which future confidence consistently improves calibration across all model families. Although transfer is naturally more challenging than within-dataset prediction, the calibration improvements remain remarkably consistent across model families, suggesting that distilled future confidence captures transferable metacognitive representations rather than dataset-specific decision boundaries.

\begin{table*}[t]
\centering
\footnotesize
\caption{Cross-dataset transfer performance across model families. Models are trained on one dataset and evaluated on a different dataset from the same reasoning domain. Lower ECE is better. Results are averaged across all three domains.}
\label{tab:model_transfer}

\setlength{\tabcolsep}{7pt}
\begin{tabular}{lcccc}
\toprule
& \multicolumn{2}{c}{25 Training Examples} &
\multicolumn{2}{c}{500 Training Examples}\\
\cmidrule(lr){2-3}\cmidrule(lr){4-5}

Model
& Pre
& Distilled
& Pre
& Distilled\\

\midrule

Llama-3.1-8B
&0.414&0.382
&0.305&0.280\\

Qwen3-7B
&0.408&0.410
&0.344&0.310\\

Mistral-Nemo-12B
&0.421&0.398
&0.285&0.252\\

Qwen3-14B
&0.392&0.414
&0.307&0.265\\

Qwen3-32B
&0.400&0.384
&0.289&0.256\\

\bottomrule
\end{tabular}
\end{table*}

\begin{table*}[t]
\centering
\footnotesize
\caption{Cross-domain transfer averaged across all models and transfer pairs. Lower ECE is better. Unlike within-domain transfer, no consistent calibration improvement is observed.}
\label{tab:cross_domain}

\begin{tabular}{ccc}
\toprule
Training Examples & Pre ECE & Distilled ECE\\
\midrule
25  & 0.567 & 0.603\\
50  & 0.598 & \textbf{0.592}\\
100 & 0.601 & \textbf{0.587}\\
200 & \textbf{0.551} & 0.592\\
300 & \textbf{0.621} & 0.632\\
400 & \textbf{0.582} & 0.621\\
500 & 0.601 & \textbf{0.561}\\
\bottomrule
\end{tabular}
\end{table*}

\subsection{Cross-Domain Transfer (Negative Result)}
\label{app:cross-domain}

Finally, we investigated whether future confidence-related information encoded in hidden representations generalises across domains by training on one domain and evaluating on another without retraining. Unlike the within-domain transfer results reported in the main paper, calibration improvements were highly inconsistent and varied across supervision budgets and model families.

Table~\ref{tab:cross_domain} shows aggregate results averaged across all models and cross-domain transfer pairs. While isolated improvements are observed under some training budgets, no stable trend emerges, and future confidence does not consistently outperform directly supervised pre-solution probes.

These findings suggest that future confidence-related information in hidden representations is transferable across datasets that share similar reasoning characteristics, but do not appear to form a universal confidence representation that generalises across fundamentally different reasoning domains such as factual recall, logical reasoning, and mathematics. We therefore restrict our transfer claims to the within-domain setting throughout the paper.

\section{Limitations and Scope}
\label{app:limits}
Our study focuses on confidence estimation in text-only instruction-tuned large language models across three representative reasoning domains: factual recall, logical reasoning, and mathematical reasoning. While these settings span substantially different cognitive demands, they do not cover multimodal reasoning, long-horizon planning, tool use, or interactive agentic workflows. Extending future confidence distillation to such settings remains an important direction for future work.

Representation-based analyses are limited to open-source models because hidden representations are not accessible through proprietary APIs. Consequently, experiments involving hidden-state probes and future confidence distillation are performed only on open-source models, whereas proprietary models are evaluated exclusively through verbal confidence estimates. As representation access becomes available for additional models, future work can investigate whether the observed confidence dynamics generalise to frontier closed-source systems.

Throughout this work, we employ linear probes and a ridge regression-based distillation objective. This choice prioritises interpretability, computational efficiency, and consistency with prior representation-probing literature, while also providing the strongest empirical performance among the architectures evaluated in our experiments. Although richer non-linear predictors may recover additional confidence-related information, they may also reduce interpretability and were therefore not the primary focus of this study.

Finally, future confidence distillation requires post-solution teacher confidence estimates during training. This supervision is obtained offline and is not required during deployment, where distilled predictors operate exclusively on pre-solution hidden representations. Our objective is therefore not to eliminate the need for post-solution confidence estimation entirely, but to amortise its computational cost by learning confidence predictors that can be applied efficiently at inference time.



\end{document}